\documentclass{article}
\usepackage{iclr2027_conference,times}
\newif\ifpreprint \preprinttrue
\ifpreprint\iclrfinalcopy\fi

\usepackage{amsmath,amsfonts,bm}

\def\eqref#1{equation~\ref{#1}}

\def\1{\bm{1}}

\DeclareMathAlphabet{\mathsfit}{\encodingdefault}{\sfdefault}{m}{sl}
\SetMathAlphabet{\mathsfit}{bold}{\encodingdefault}{\sfdefault}{bx}{n}

\usepackage{hyperref}
\usepackage{url}
\usepackage{graphicx}
\usepackage{booktabs}
\usepackage{amsmath,amssymb}
\usepackage{multirow}
\usepackage{adjustbox}
\usepackage{placeins}
\usepackage{float}
\usepackage{xcolor}
\usepackage{colortbl}
\definecolor{oursrow}{RGB}{235,243,252}

\usepackage[ruled,vlined]{algorithm2e}

\newcommand{\ours}{OmniHSR}
\title{Super-Resolving Unseen Hyperspectral\\Sensors at Any Scale via Spatial Operators}

\author{Ji-Xuan He\textsuperscript{1}\quad Guohang Zhuang\textsuperscript{2}\quad Bo Junge\textsuperscript{1}\quad Tingyi Li\textsuperscript{2}\quad Lingchen\textsuperscript{1}\\
\bf Miaomiao Cai\textsuperscript{3}\quad Yanan Qiao\textsuperscript{1,\textdagger}\quad Xiujin Liu\textsuperscript{4}\quad Junfeng Fang\textsuperscript{3}\\
\textsuperscript{1}Xi'an Jiaotong University\quad \textsuperscript{2}Hefei University of Technology\\
\textsuperscript{3}National University of Singapore\quad \textsuperscript{4}University of Michigan\\
\textsuperscript{\textdagger}Corresponding author.
}

\begin{document}

\maketitle
\ifpreprint\lhead{Preprint}\vspace{-24pt}\fi

\begin{figure}[H]
\centering
\includegraphics[width=\linewidth]{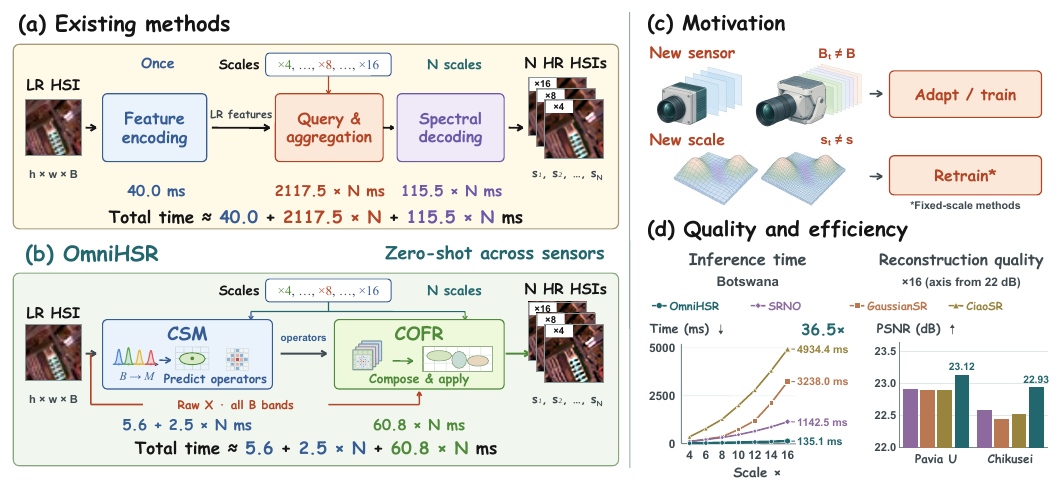}
\caption{\textbf{Cross-sensor, arbitrary-scale reconstruction with \ours{}.}
(a) Existing methods, here CiaoSR, decode the spectral values of every band from features and repeat querying and decoding per requested scale.
(b) \ours{} predicts local spatial operators with CSM and applies them to all observed bands with COFR.
(c) A new sensor requires adapting or retraining existing methods, and a new scale requires retraining fixed-scale methods.
(d) At $\times16$, \ours{} reaches a higher zero-shot PSNR than the baselines on Pavia~U and Chikusei and runs $36.5\times$ faster than CiaoSR on the $145$-band Botswana.
$N$ is the number of requested scales; times are measured on Botswana.}
\label{fig:teaser}
\end{figure}

\begin{abstract}
Achieving cross-sensor generalization and arbitrary-scale reconstruction with a single model remains challenging in hyperspectral super-resolution (HSR). Although recent methods support arbitrary-scale reconstruction, applying them to new sensors or scales beyond the training range often requires additional data and computation to maintain reconstruction quality. To address these challenges, we propose OmniHSR, which predicts band-shared spatial operators rather than spectral values. Cross-Spectral Mapping (CSM) resamples inputs with any number of bands to fixed reference positions and predicts local operators with Gaussian supports. Continuous Operator-Field Reconstruction (COFR) composes these operators into a continuous field and applies them to all original bands for arbitrary-scale reconstruction. Experiments demonstrate that operator prediction outperforms direct spectral-value prediction on all seven datasets. Trained solely on ARAD with only 0.538M parameters, OmniHSR outperforms all directly transferred baselines on six unseen datasets without target-domain training data or adaptation. Across twelve upsampling factors from $\times2$ to $\times48$, it improves average PSNR on Pavia U and Chikusei by 0.55 dB over the strongest baseline. It also surpasses baselines trained from scratch or adapted on the target sensor and achieves up to $36\times$ faster inference. Our code will be publicly released soon.

\end{abstract}

\section{Introduction}
\label{sec:intro}

Hyperspectral images provide rich spectral information but often lack sufficient spatial detail for practical applications, motivating hyperspectral image super-resolution (HSR)~\citep{bian2024hyperspectral,hyperfree,ciotola2025,guo2023}. Most existing HSR methods are designed for fixed upsampling factors such as $\times2$, $\times4$, and $\times8$~\citep{gelin,essaformer,cst,hsrmamba}, while recent arbitrary-scale methods handle different factors with a single model~\citep{liif,gaussiansr,sqformer,spgassr}. Hyperspectral sensors also differ in their band configurations, from $31$ to more than one hundred bands, and the public data of a remote-sensing sensor is often a single scene. Achieving cross-sensor generalization and arbitrary-scale reconstruction with a single model remains challenging.

Existing HSR networks output the spectral values of every band, with as many output channels as the training sensor has bands (Figure~\ref{fig:teaser}a). To serve a new sensor, they either run on sliding windows of bands, which multiplies the inference cost with the band count, or are trained from scratch or adapted on the target sensor~\citep{slr,deflect}, which needs target data (Figure~\ref{fig:teaser}c). Beyond their training range of factors, fixed-scale methods must be retrained and arbitrary-scale methods lose accuracy. In all these cases, the network itself still predicts the spectral values, for every band and at every factor.

To address these challenges, we propose \ours{}, which predicts band-shared spatial operators instead of spectral values (Figure~\ref{fig:teaser}b). Every pixel of a low-resolution observation is a spatial mixture of the high-resolution spectra within its footprint, and all bands are mixed in the same way, so a high-resolution spectrum at any position can be approximated by combining the observed spectra around it, and the network only has to decide how they are combined. Such an operator carries no band dimension, so it acts on all bands of any sensor, and the network runs once regardless of the band count. Networks that predict filters for their input already work this way in denoising and video super-resolution~\citep{kpn,duf}, for a fixed set of channels on a fixed output grid.

Predicting such operators for any sensor at any factor poses two difficulties. The network needs an input of fixed size while the band count changes with the sensor, and the operators are predicted on the low-resolution grid but must act on a target grid at any factor. Cross-Spectral Mapping (CSM) resamples an input with any number of bands to fixed reference positions and predicts, for every low-resolution pixel, a local operator with a Gaussian support~\citep{kerbl3dgs,gaussiansr,gsasr}. Continuous Operator-Field Reconstruction (COFR) composes these operators along their supports into a continuous field, evaluates it on the target grid at any factor, and applies it to all original bands.

With the backbone and training shared, predicting operators is more accurate than predicting spectral values on all seven datasets. Trained solely on the $31$-band ARAD~\citep{arad2022ntire} with only $0.538$M parameters, \ours{} outperforms all directly transferred baselines on six unseen datasets without target-domain training data or adaptation, and across twelve factors from $\times2$ to $\times48$ it improves the average PSNR on Pavia~U~\citep{pavia} and Chikusei~\citep{chikusei} by $0.55$~dB over the strongest baseline. It also surpasses the baselines trained from scratch or adapted on the target sensor and runs up to $36\times$ faster on sensors with more than $31$ bands (Figure~\ref{fig:teaser}d). Our main contributions are as follows.

\begin{itemize}
\item We propose \ours{}, a cross-sensor arbitrary-scale HSR framework that predicts band-shared spatial operators instead of spectral values, so that one model trained once serves unseen sensors at any factor zero-shot.
\item We propose CSM, which maps an input with any number of bands to fixed reference positions and predicts local operators with Gaussian supports, and COFR, which composes these operators into a continuous field and applies it to all original bands for arbitrary-scale reconstruction.
\item Experiments on seven datasets from six sensors show that, with the backbone and training shared, predicting operators is more accurate than predicting spectral values on every dataset, and that the zero-shot \ours{} outperforms the baselines whether they are transferred directly, trained from scratch or adapted on the target sensor, and runs up to $36\times$ faster.
\end{itemize}

\section{Related Work}
\label{sec:related}

\paragraph{Hyperspectral image super-resolution.} Single-image hyperspectral super-resolution has moved from grouped convolutional networks to attention, transformer and state-space designs~\citep{gelin,essaformer,cst,hsrmamba}. These networks are trained on one sensor at a time, and their last layer has as many channels as that sensor has bands. SQformer, SPG-ASSR and later methods~\citep{sqformer,spgassr,mcarb,dcmarb} extend single-image hyperspectral super-resolution to arbitrary factors. For varying band counts, denoisers that run 3D convolutions or quasi-recurrent units along the spectral axis accept any band count, and models trained on 31-band natural scenes run on sensors with more bands~\citep{qrnn3d,hsdt}. EigenSR~\citep{eigensr} super-resolves the spatial eigenimages of a hyperspectral image with a pretrained RGB network, and MLSR~\citep{mlsr} is guided by an RGB image and handles band sets sampled from 31-band datasets, with one model per factor. Remote-sensing foundation models accept different sets of channels through spatial-spectral tokens, wavelength-conditioned embeddings, hypernetworks or attention across channels for recognition and interpretation~\citep{spectralgpt,hyperfree,copernicusfm,carl}. Fusion methods reach arbitrary factors or band counts with a high-resolution guide image~\citep{arbrpn,ssa,sfno,clorf}, and high-pass filtering in pansharpening adds the high-pass detail of a panchromatic image to the resampled multispectral bands~\citep{chavez1991comparison}. \ours{} needs no wavelength, guide image or adaptation, covers any factor with one model, transfers from 31 to up to 145 bands, and its network outputs no spectral value.

\paragraph{Arbitrary-scale super-resolution and predicted filters.} Meta-SR~\citep{metasr} and ArbSR~\citep{arbsr} condition the upsampling layer on the factor, LIIF~\citep{liif} queries a local implicit function at continuous coordinates, and later work refines the query with texture estimation, learned ensembles, latent modulation, neural operators, normalizing flows and diffusion~\citep{lte,ciaosr,lmf,srno,linf,idm,neuropdiff}; CUF~\citep{cuf} learns continuous upsampling kernels inside such a network. Gaussian splatting represents a scene or an image with primitives of continuous spatial support~\citep{kerbl3dgs,gaussianimage}, and GaussianSR, GSASR and ContinuousSR~\citep{gaussiansr,gsasr,continuoussr} predict 2D Gaussians from the low-resolution image and render them onto any target grid. These methods decode the value at a query position from network features into a number of channels fixed at training time. Predicting a filter and applying it to the input is established in dynamic filtering, burst denoising, video super-resolution and feature upsampling~\citep{dfn,kpn,duf,carafe}. In joint image filtering, DKN~\citep{dkn} makes each predicted kernel sum to zero, applies it to the bicubic-upsampled target and adds the result back, and it runs separately on each channel of a multi-channel target. LeRF~\citep{lerf} interpolates the input image with an anisotropic Gaussian predicted for every input pixel, and \citet{wronski2019handheld} merge a burst by splatting anisotropic kernels onto an arbitrary output grid. \ours{} keeps the continuous support of the Gaussian primitives and lets every primitive carry one zero-sum operator, which is predicted from the low-resolution image alone and acts on every band of an unseen sensor following the band-shared prior.

\section{Proposed Method}
\label{sec:method}

Cross-sensor arbitrary-scale HSR reconstructs a high-resolution hyperspectral image $\hat{Y} \in \mathbb{R}^{sh \times sw \times B}$ from a low-resolution observation $X \in \mathbb{R}^{h \times w \times B}$, where the band number $B$ varies across sensors and the scale factor $s$ is arbitrary. \ours{} aims to build a unified model that generalizes across different band configurations and generates outputs at arbitrary scales.

\begin{figure}[t]
\centering
\includegraphics[width=\linewidth]{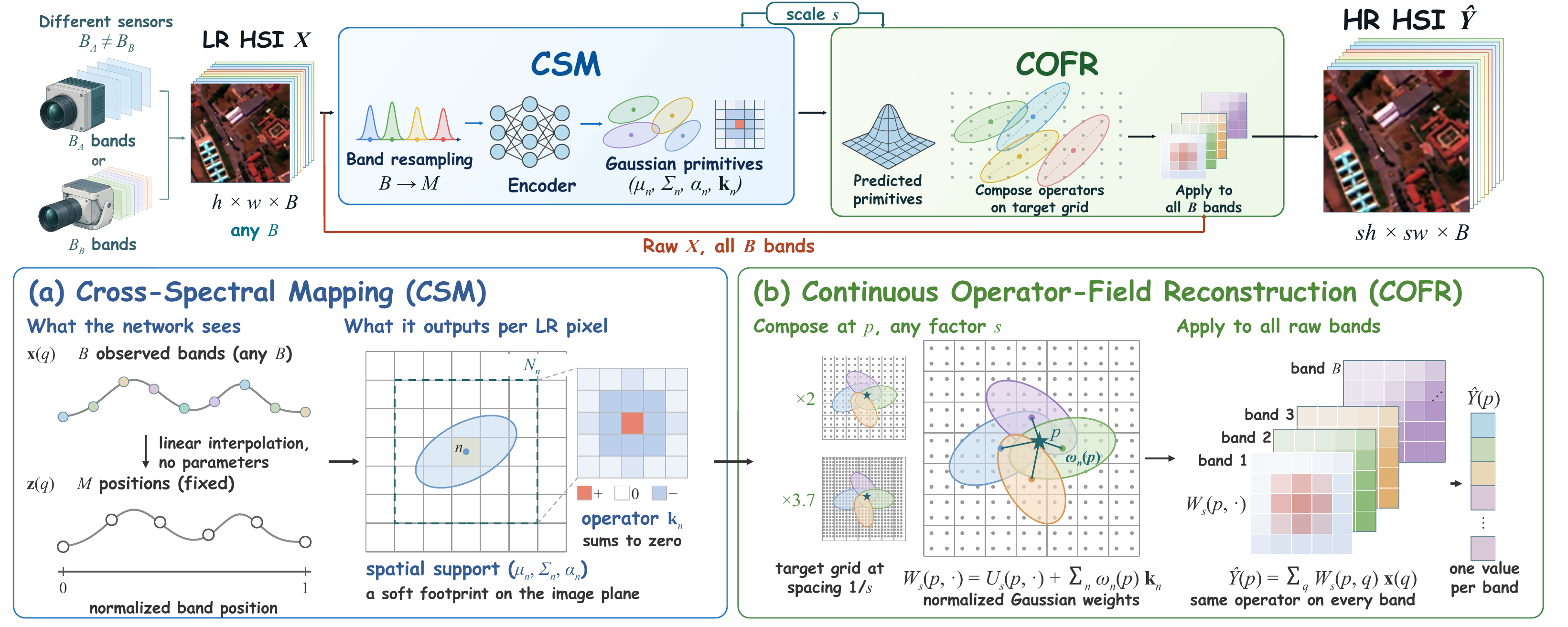}
\caption{Overview of \ours. (a) CSM resamples $B$ input bands to $M$ fixed normalized band positions and predicts scale-conditioned spatial supports and zero-sum operators. (b) COFR composes the operators with normalized Gaussian weights into one operator at target positions spaced $1/s$ apart and applies it to all raw $B$ bands.}
\label{fig:framework}
\vspace{-6pt}
\end{figure}

\ours{} consists of two modules, Cross-Spectral Mapping (CSM) and Continuous Operator-Field Reconstruction (COFR), and its overall pipeline is illustrated in Figure~\ref{fig:framework}. Specifically,
First, CSM resamples $X$ along the band axis to a fixed number $M$ of bands, feeds the result to an encoder and predicts $hw$ 2D Gaussian primitives conditioned on the factor $s$. Each primitive has a Gaussian spatial support, given by its center $\mu_n$, covariance $\Sigma_n$ and opacity $\alpha_n$, and a local spatial operator $\mathbf{k}_n$, which is a small spatial filter,
\begin{equation}
\label{eq:overview:csm}
\{X,\, s\} \xrightarrow{\;\text{CSM}\;}
\bigl\{(\mu_n, \Sigma_n, \alpha_n, \mathbf{k}_n)\bigr\}_{n=1}^{hw} .
\end{equation}
These primitives carry no band axis, and the $M$ bands appear only inside the network. Then, COFR composes the operators of all primitives according to their spatial supports into one reconstruction operator on the $sh \times sw$ target grid and applies it to the raw input $X$, with all $B$ bands sharing the same coefficients, which gives $\hat{Y}$,
\begin{equation}
\label{eq:overview:cofr}
\left(
\bigl\{(\mu_n, \Sigma_n, \alpha_n, \mathbf{k}_n)\bigr\}_{n=1}^{hw},
X,\,
s
\right)
\xrightarrow{\;\text{COFR}\;}
\hat{Y}.
\end{equation}

\subsection{Band-Shared Operator Prior}
\label{sec:method:prior}

\textbf{Observation and band-shared operator prior.} Let $Y \in \mathbb{R}^{sh \times sw \times B}$ be the high-resolution image, and let the observation $X$ be obtained by applying the same spatial degradation to every band. The bicubic upsampling $U_s X$ combines the observed pixels with fixed weights that sum to one and recovers only the smooth part of $Y$. Within a local window, we let a zero-sum $5\times5$ operator $\mathbf{k}$ change how the observed pixels are combined,
\begin{equation}
\label{eq:prior}
Y_b \approx U_s \bigl((\delta + \mathbf{k}) * X_b\bigr),
\qquad
\sum_{q} k_q = 0,
\end{equation}
where $b$ indexes the bands, $*$ applies the operator of each window on the low-resolution grid and $\delta$ is the identity operator, so the combination $\delta + \mathbf{k}$ keeps weights that sum to one and reduces to bicubic interpolation when $\mathbf{k} = 0$. When all bands undergo the same spatial degradation, how the pixels in a window are combined is decided by the same spatial structure, so one $\mathbf{k}$ should serve all bands, which we call the band-shared operator prior.

This prior decides what the network should output. One operator serves all bands, so the network only needs to output one operator without a band dimension for every primitive. This operator can be determined from a fixed set of bands, so the network input only needs a fixed set of band positions. CSM realizes these two points (Section~\ref{sec:method:csm}). The values differ from band to band, and COFR takes them directly from the raw observation while it places the operators on the target grid at any factor (Section~\ref{sec:method:cofr}).

\subsection{Cross-Spectral Mapping}
\label{sec:method:csm}

\textbf{Input side.} Let the $B$ bands of $X$ be ordered by increasing wavelength, with the normalized position of band $b$ written as $(b-1)/(B-1)$. CSM interpolates the spectrum of every pixel linearly along the band axis onto $M$ equally spaced reference positions in $[0,1]$, which gives the network input $Z$, then extracts features with an encoder $E_\theta$ and modulates them by the factor $s$,
\begin{equation}
\label{eq:canon}
\mathbf{z}(q) = C_B\, \mathbf{x}(q),
\qquad
F_s = \gamma(s) \odot E_\theta(Z) + \beta(s),
\end{equation}
where $\mathbf{x}(q) \in \mathbb{R}^{B}$ and $\mathbf{z}(q) \in \mathbb{R}^{M}$ are the spectra of $X$ and $Z$ at low-resolution pixel $q$, $C_B \in \mathbb{R}^{M \times B}$ is the matrix of this linear interpolation, which depends only on $B$ and has no learnable parameter, and $M = 31$ in this paper; $\gamma(s)$ and $\beta(s)$ are produced from $\log_2 s$ by a two-layer perceptron.

These reference positions record only the order of the bands, so sensors with different spectral ranges share one input axis without wavelength values. Band order is enough because $Z$ serves only to predict spatial operators, which are decided by the spatial structure shared by all bands (Section~\ref{sec:method:prior}).

\textbf{Output side.} From $F_s$, CSM predicts one 2D Gaussian primitive at every low-resolution pixel, $hw$ primitives in total, and the pixel of primitive $n$ is written as $q_n$. A primitive is described by a center $\mu_n$, a covariance $\Sigma_n$, an opacity $\alpha_n$ and a $5\times5$ local spatial operator $\mathbf{k}_n$,
\begin{equation}
\label{eq:primitive}
\bigl(\mu_n,\; \Sigma_n,\; \alpha_n,\; \mathbf{k}_n\bigr),
\qquad
\mu_n \in \mathbb{R}^{2},\quad \Sigma_n \in \mathbb{R}^{2\times2},\quad
\alpha_n \in (0, 1),\quad \mathbf{k}_n \in \mathbb{R}^{25}.
\end{equation}
The first three give the spatial support of the primitive, and $\mathbf{k}_n$ is the operator that this primitive will apply to $X$. The prediction head outputs $25$ coefficients, and their mean is subtracted so that they sum to zero,
\begin{equation}
\label{eq:zerosum}
\sum_{q \in \mathcal{N}_n} k_{n,q} = 0,
\end{equation}
where $\mathcal{N}_n$ is the $5\times5$ window of low-resolution pixels centered at $q_n$ and $k_{n,q}$ is the entry of $\mathbf{k}_n$ at $q$. With the zero sum, the weights with which the observed pixels are combined still sum to one, so a region whose pixels are all equal stays unchanged. No parameter of the network is tied to $B$, so one set of weights serves any band count.

\subsection{Continuous Operator-Field Reconstruction}
\label{sec:method:cofr}

Eq.~(\ref{eq:prior}) places the operator of a window on the target grid with bicubic weights, which are fixed and symmetric around every low-resolution pixel. COFR lets every primitive carry its own operator and places it with the anisotropic Gaussian support of the primitive, which can stretch along the edges that run between low-resolution pixels and can be evaluated at any continuous position.

\begin{table}[t]
\centering
\caption{PSNR$\uparrow$ (dB) at twelve factors on the in-domain ARAD and on two unseen sensors. \textbf{Bold}: best; \underline{underline}: second best; gray: below bicubic interpolation. Time: mean time per image.}
\label{tab:main}
\footnotesize\renewcommand{\arraystretch}{1.12}\setlength{\tabcolsep}{2.6pt}
\begin{adjustbox}{max width=\linewidth,center}
\begin{tabular}{@{}cc*{12}{c}c@{}}
\toprule
\multirow{2}{*}{Dataset} & \multirow{2}{*}{Method} & \multicolumn{3}{c}{In-range integers} & \multicolumn{4}{c}{In-range fractional} & \multicolumn{5}{c}{Out-of-range extrapolation} & \multirow{2}{*}{\begin{tabular}{@{}c@{}}Time\\(ms)\end{tabular}} \\
\cmidrule(lr){3-5}\cmidrule(lr){6-9}\cmidrule(lr){10-14}
 & & $\times$2 & $\times$4 & $\times$8 & $\times$2.5 & $\times$3.7 & $\times$5.3 & $\times$7.6 & $\times$12 & $\times$16 & $\times$24 & $\times$36 & $\times$48 &  \\
\midrule
\multirow{8}{*}{\rotatebox{90}{\begin{tabular}{@{}c@{}}ARAD\\($31$ bands)\end{tabular}}} & Bicubic & 44.04 & 37.50 & 33.37 & 41.60 & 38.08 & 35.61 & 33.70 & 31.52 & 30.50 & 29.20 & \underline{27.94} & \underline{27.05} & -- \\
 & LIIF & 44.48 & 37.96 & 33.56 & 42.15 & 38.56 & 35.99 & 33.90 & \textcolor{gray}{31.47} & \textcolor{gray}{30.31} & \textcolor{gray}{28.89} & \textcolor{gray}{27.55} & \textcolor{gray}{26.65} & 16.5 \\
 & LTE & 45.33 & 38.38 & 33.87 & 42.83 & 39.01 & 36.34 & 34.23 & 31.79 & 30.65 & 29.22 & \textcolor{gray}{27.79} & \textcolor{gray}{26.87} & 17.6 \\
 & Meta-SR & 45.42 & 38.33 & 33.81 & 42.79 & 38.94 & 36.27 & 34.16 & 31.73 & 30.60 & \textcolor{gray}{29.17} & \textcolor{gray}{27.75} & \textcolor{gray}{26.84} & 51.1 \\
 & CiaoSR & 45.39 & 38.39 & 33.86 & 42.80 & 39.00 & 36.34 & 34.21 & 31.75 & 30.59 & \textcolor{gray}{29.12} & \textcolor{gray}{27.66} & \textcolor{gray}{26.71} & 57.2 \\
 & SRNO & \underline{45.58} & \underline{38.49} & \underline{33.92} & \underline{42.94} & \underline{39.10} & \underline{36.41} & \underline{34.28} & \underline{31.83} & \underline{30.67} & \underline{29.26} & \textcolor{gray}{27.82} & \textcolor{gray}{26.76} & 15.6 \\
 & GaussianSR & 44.71 & 38.23 & 33.72 & 42.46 & 38.81 & 36.19 & 34.08 & 31.62 & \textcolor{gray}{30.47} & \textcolor{gray}{28.99} & \textcolor{gray}{27.49} & \textcolor{gray}{26.47} & 12.5 \\
 & \cellcolor{oursrow}\ours{} & \cellcolor{oursrow}\textbf{48.41} & \cellcolor{oursrow}\textbf{40.33} & \cellcolor{oursrow}\textbf{35.24} & \cellcolor{oursrow}\textbf{45.34} & \cellcolor{oursrow}\textbf{41.03} & \cellcolor{oursrow}\textbf{38.00} & \cellcolor{oursrow}\textbf{35.63} & \cellcolor{oursrow}\textbf{33.02} & \cellcolor{oursrow}\textbf{31.71} & \cellcolor{oursrow}\textbf{30.09} & \cellcolor{oursrow}\textbf{28.43} & \cellcolor{oursrow}\textbf{27.42} & \cellcolor{oursrow}\textbf{11.5} \\
\midrule
\multirow{8}{*}{\rotatebox{90}{\begin{tabular}{@{}c@{}}Pavia~U\\($103$ bands)\end{tabular}}} & Bicubic & 32.64 & 28.13 & 25.04 & 30.92 & 28.51 & 26.78 & 25.32 & \underline{23.51} & \underline{22.93} & \underline{22.41} & \underline{21.89} & \underline{21.54} & -- \\
 & LIIF & \textcolor{gray}{32.62} & \textcolor{gray}{28.11} & \textcolor{gray}{24.93} & 30.96 & 28.52 & \textcolor{gray}{26.74} & \textcolor{gray}{25.20} & \textcolor{gray}{23.39} & \textcolor{gray}{22.81} & \textcolor{gray}{22.28} & \textcolor{gray}{21.75} & \textcolor{gray}{21.43} & 115.4 \\
 & LTE & 33.03 & 28.34 & 25.13 & 31.28 & 28.75 & 26.93 & 25.39 & \textcolor{gray}{23.47} & \textcolor{gray}{22.91} & \textcolor{gray}{22.37} & \textcolor{gray}{21.84} & \textcolor{gray}{21.47} & 120.2 \\
 & Meta-SR & 33.14 & 28.35 & 25.07 & 31.30 & 28.76 & 26.91 & 25.33 & \textcolor{gray}{23.45} & \textcolor{gray}{22.89} & \textcolor{gray}{22.34} & \textcolor{gray}{21.80} & \textcolor{gray}{21.43} & 357.4 \\
 & CiaoSR & 33.03 & 28.35 & 25.12 & 31.25 & 28.75 & 26.94 & 25.38 & \textcolor{gray}{23.48} & \textcolor{gray}{22.88} & \textcolor{gray}{22.30} & \textcolor{gray}{21.71} & \textcolor{gray}{21.39} & 398.3 \\
 & SRNO & \underline{33.24} & \underline{28.44} & \underline{25.19} & \underline{31.37} & \underline{28.83} & \underline{27.01} & \underline{25.45} & \textcolor{gray}{23.49} & \textcolor{gray}{22.91} & \textcolor{gray}{22.37} & \textcolor{gray}{21.84} & \textcolor{gray}{21.34} & 106.9 \\
 & GaussianSR & 32.88 & 28.31 & 25.07 & 31.18 & 28.71 & 26.87 & 25.33 & \textcolor{gray}{23.47} & \textcolor{gray}{22.89} & \textcolor{gray}{22.35} & \textcolor{gray}{21.80} & \textcolor{gray}{21.43} & 85.0 \\
 & \cellcolor{oursrow}\ours{} & \cellcolor{oursrow}\textbf{34.29} & \cellcolor{oursrow}\textbf{28.94} & \cellcolor{oursrow}\textbf{25.62} & \cellcolor{oursrow}\textbf{32.17} & \cellcolor{oursrow}\textbf{29.33} & \cellcolor{oursrow}\textbf{27.46} & \cellcolor{oursrow}\textbf{25.90} & \cellcolor{oursrow}\textbf{23.84} & \cellcolor{oursrow}\textbf{23.12} & \cellcolor{oursrow}\textbf{22.57} & \cellcolor{oursrow}\textbf{21.98} & \cellcolor{oursrow}\textbf{21.57} & \cellcolor{oursrow}\textbf{20.5} \\
\midrule
\multirow{8}{*}{\rotatebox{90}{\begin{tabular}{@{}c@{}}Chikusei\\($128$ bands)\end{tabular}}} & Bicubic & 33.04 & 27.60 & 24.50 & 31.17 & 28.07 & 26.05 & 24.70 & 23.26 & 22.53 & \underline{21.78} & \textbf{21.10} & \underline{20.66} & -- \\
 & LIIF & \textcolor{gray}{32.82} & \textcolor{gray}{27.52} & \textcolor{gray}{24.38} & \textcolor{gray}{31.09} & \textcolor{gray}{28.02} & \textcolor{gray}{25.98} & \textcolor{gray}{24.61} & \textcolor{gray}{23.12} & \textcolor{gray}{22.38} & \textcolor{gray}{21.60} & \textcolor{gray}{20.93} & \textcolor{gray}{20.53} & 165.3 \\
 & LTE & 33.65 & 27.89 & 24.58 & 31.71 & 28.42 & 26.26 & 24.82 & 23.32 & 22.55 & \textcolor{gray}{21.73} & \textcolor{gray}{21.01} & \textcolor{gray}{20.57} & 171.4 \\
 & Meta-SR & 33.76 & 27.85 & 24.54 & 31.71 & 28.38 & 26.21 & 24.77 & 23.26 & \textcolor{gray}{22.50} & \textcolor{gray}{21.68} & \textcolor{gray}{20.98} & \textcolor{gray}{20.56} & 512.3 \\
 & CiaoSR & 33.68 & 27.88 & 24.58 & 31.68 & 28.39 & 26.24 & 24.81 & 23.29 & \textcolor{gray}{22.52} & \textcolor{gray}{21.66} & \textcolor{gray}{20.90} & \textcolor{gray}{20.50} & 570.0 \\
 & SRNO & \underline{33.81} & \underline{27.96} & \underline{24.62} & \underline{31.81} & \underline{28.48} & \underline{26.31} & \underline{24.85} & \underline{23.35} & \underline{22.57} & \textcolor{gray}{21.74} & \textcolor{gray}{\underline{21.03}} & \textcolor{gray}{20.58} & 152.4 \\
 & GaussianSR & 33.31 & 27.76 & \textcolor{gray}{24.48} & 31.48 & 28.27 & 26.13 & \textcolor{gray}{24.70} & \textcolor{gray}{23.20} & \textcolor{gray}{22.44} & \textcolor{gray}{21.62} & \textcolor{gray}{20.90} & \textcolor{gray}{20.47} & 121.3 \\
 & \cellcolor{oursrow}\ours{} & \cellcolor{oursrow}\textbf{35.44} & \cellcolor{oursrow}\textbf{28.95} & \cellcolor{oursrow}\textbf{25.20} & \cellcolor{oursrow}\textbf{33.13} & \cellcolor{oursrow}\textbf{29.51} & \cellcolor{oursrow}\textbf{27.10} & \cellcolor{oursrow}\textbf{25.47} & \cellcolor{oursrow}\textbf{23.78} & \cellcolor{oursrow}\textbf{22.93} & \cellcolor{oursrow}\textbf{21.92} & \cellcolor{oursrow}\textbf{21.10} & \cellcolor{oursrow}\textbf{20.67} & \cellcolor{oursrow}\textbf{27.1} \\
\bottomrule
\end{tabular}
\end{adjustbox}
\vspace{-6pt}
\end{table}

\textbf{Composing onto any target grid.} All positions are measured in units of low-resolution pixels, so the target grid at factor $s$ has a pixel spacing of $1/s$. At a continuous position $p$, the response of primitive $n$ is given by an anisotropic Gaussian, and normalizing it by the total response of all primitives gives the weight $\omega_n(p)$,
\begin{equation}
\label{eq:render}
G_n(p) = \exp\!\Bigl(-\tfrac{1}{2}\,(p - \mu_n)^{\top} \Sigma_n^{-1} (p - \mu_n)\Bigr),
\qquad
\omega_n(p) = \frac{\alpha_n G_n(p)}{\sum_j \alpha_j G_j(p) + \epsilon},
\end{equation}
where $\epsilon$ is a small constant.

\textbf{Continuous operator field.} At a continuous position $p$, mixing the operators of the primitives around $p$ with these weights gives one local operator on the observed pixels,
\begin{equation}
\label{eq:effective}
A_s(p, q) = \sum_{n \,:\, q \in \mathcal{N}_n} \omega_n(p)\, k_{n,q}.
\end{equation}
Each coefficient $A_s(p, q)$ is a scalar that does not depend on the band. In this way, the discrete operators $\mathbf{k}_n$ extend to a continuous operator field $A_s$ over the whole image plane, and the target grid at factor $s$ is one sampling of $A_s$.

\textbf{Reconstruction as an operator.} COFR adds the operator field to the bicubic weights, which sum to one, and applies the result to all bands of $X$. With the spatial positions flattened, the reconstruction reads
\begin{equation}
\label{eq:operator}
\hat{Y} = W_s X,
\qquad
W_s = U_s + R_{Z,s} K_{Z,s},
\qquad
W_s \mathbf{1} = \mathbf{1},
\end{equation}
where $X$ and $\hat{Y}$ are flattened into $hw \times B$ and $s^2hw \times B$ matrices, $U_s$ is the bicubic upsampling matrix at factor $s$, and $\mathbf{1}$ is the all-ones vector. Row $n$ of $K_{Z,s}$ holds $\mathbf{k}_n$ placed on the window $\mathcal{N}_n$ and $R_{Z,s}$ collects the weights $\omega_n(p)$; both are predicted from $Z$ and $s$, and the entries of $R_{Z,s} K_{Z,s}$ are the coefficients $A_s(p, q)$ of Eq.~(\ref{eq:effective}). Every row of $U_s$ sums to one and every row of $K_{Z,s}$ sums to zero, so every row of $W_s$ sums to one, and $W_s$ reduces to bicubic interpolation when every operator is zero. Row $p$ of Eq.~(\ref{eq:operator}), drawn in Figure~\ref{fig:framework}b, reads $\hat{Y}(p) = \sum_q W_s(p, q)\, \mathbf{x}(q)$ with $W_s(p, q) = U_s(p, q) + A_s(p, q)$, so the network decides only how the observed pixels are combined, and every spectrum of $\hat{Y}$ is an affine combination of the spectra of $X$.

\textbf{What the operators add.} Applied to $X$, the operator field adds to the bicubic reconstruction at $p$
\begin{equation}
\label{eq:anchor}
\sum_q A_s(p, q)\, \mathbf{x}(q) = \sum_n \omega_n(p)\, \mathbf{r}_n,
\qquad
\mathbf{r}_n = \sum_{q \in \mathcal{N}_n} k_{n,q}\, \mathbf{x}(q)
= \sum_{q \in \mathcal{N}_n} k_{n,q} \bigl[\, \mathbf{x}(q) - \mathbf{x}(q_n) \,\bigr],
\end{equation}
where the first equality exchanges the order of summation in Eq.~(\ref{eq:effective}) and the last follows from Eq.~(\ref{eq:zerosum}), so $\mathbf{r}_n$ consists entirely of spectral differences to the center pixel and is computed without any parameter that depends on $B$.

For a given $W_s$, the reconstruction commutes with any affine spectral map applied to every pixel. For any $P \in \mathbb{R}^{B' \times B}$ and $\mathbf{c} \in \mathbb{R}^{B'}$,
\begin{equation}
\label{eq:equivariance}
W_s \bigl( X P^{\top} + \mathbf{1} \mathbf{c}^{\top} \bigr)
= \bigl( W_s X \bigr) P^{\top} + \mathbf{1} \mathbf{c}^{\top},
\end{equation}
where the linear part holds because $W_s$ acts only on the spatial axes, and the constant part holds because every row of $W_s$ sums to one. Taking a subset of the bands, merging neighboring bands and applying a per-band gain and offset are all such maps, and for a given $W_s$ they pass through the reconstruction exactly, so the band definition of a sensor reaches the output of \ours{} only through the prediction of $W_s$ from $Z$, while it reaches every output value of a network that outputs spectral values. Section~\ref{sec:exp} tests this prediction on sensors unseen in training. Rows that sum to one are also the partition-of-unity condition of interpolation kernels, which the zero sum keeps for the learned operator field. Computationally, the encoder $E_\theta(Z)$ runs once for any number of requested factors, and the only step whose cost grows with $B$ is the product $W_s X$, which is linear in $B$.

\begin{table}[t]
\centering
\caption{SSIM$\uparrow$ and SAM$\downarrow$ (degrees) on Pavia~U at the factors of Table~\ref{tab:main}. \textbf{Bold}: best; \underline{underline}: second best; gray: worse than bicubic interpolation.}
\label{tab:ssim_sam}
\footnotesize\renewcommand{\arraystretch}{1.12}\setlength{\tabcolsep}{2.6pt}
\begin{adjustbox}{max width=\linewidth,center}
\begin{tabular}{@{}cc*{12}{c}@{}}
\toprule
\multirow{2}{*}{Metric} & \multirow{2}{*}{Method} & \multicolumn{3}{c}{In-range integers} & \multicolumn{4}{c}{In-range fractional} & \multicolumn{5}{c}{Out-of-range extrapolation} \\
\cmidrule(lr){3-5}\cmidrule(lr){6-9}\cmidrule(lr){10-14}
 & & $\times$2 & $\times$4 & $\times$8 & $\times$2.5 & $\times$3.7 & $\times$5.3 & $\times$7.6 & $\times$12 & $\times$16 & $\times$24 & $\times$36 & $\times$48 \\
\midrule
\multirow{8}{*}{\rotatebox{90}{SSIM$\uparrow$}} & Bicubic & 0.918 & 0.762 & 0.604 & 0.875 & 0.781 & 0.692 & 0.617 & 0.538 & 0.516 & 0.499 & \underline{0.489} & \textbf{0.483} \\
 & LIIF & \textcolor{gray}{0.917} & 0.764 & \textcolor{gray}{0.602} & 0.876 & 0.783 & 0.694 & \textcolor{gray}{0.616} & \textcolor{gray}{0.536} & \textcolor{gray}{0.513} & \textcolor{gray}{0.497} & \textcolor{gray}{0.486} & \textcolor{gray}{0.481} \\
 & LTE & 0.923 & 0.774 & 0.612 & 0.884 & 0.793 & 0.703 & 0.625 & \underline{0.541} & \underline{0.517} & \underline{0.500} & \textcolor{gray}{0.488} & \textcolor{gray}{\underline{0.482}} \\
 & Meta-SR & 0.926 & 0.774 & 0.610 & 0.885 & 0.793 & 0.702 & 0.623 & 0.539 & 0.516 & \textcolor{gray}{0.499} & \textcolor{gray}{0.487} & \textcolor{gray}{0.481} \\
 & CiaoSR & 0.924 & 0.773 & 0.612 & 0.883 & 0.792 & 0.703 & 0.625 & 0.540 & 0.516 & \textcolor{gray}{0.498} & \textcolor{gray}{0.486} & \textcolor{gray}{0.481} \\
 & SRNO & \underline{0.927} & \underline{0.777} & \underline{0.614} & \underline{0.886} & \underline{0.795} & \underline{0.706} & \underline{0.628} & \underline{0.541} & \underline{0.517} & \textcolor{gray}{0.499} & \textcolor{gray}{0.487} & \textcolor{gray}{0.478} \\
 & GaussianSR & 0.922 & 0.772 & 0.609 & 0.882 & 0.790 & 0.700 & 0.622 & 0.539 & \textcolor{gray}{0.516} & \textcolor{gray}{0.498} & \textcolor{gray}{0.487} & \textcolor{gray}{0.481} \\
 & \cellcolor{oursrow}\ours{} & \cellcolor{oursrow}\textbf{0.943} & \cellcolor{oursrow}\textbf{0.803} & \cellcolor{oursrow}\textbf{0.639} & \cellcolor{oursrow}\textbf{0.907} & \cellcolor{oursrow}\textbf{0.819} & \cellcolor{oursrow}\textbf{0.732} & \cellcolor{oursrow}\textbf{0.654} & \cellcolor{oursrow}\textbf{0.557} & \cellcolor{oursrow}\textbf{0.526} & \cellcolor{oursrow}\textbf{0.504} & \cellcolor{oursrow}\textbf{0.490} & \cellcolor{oursrow}\textbf{0.483} \\
\midrule
\multirow{8}{*}{\rotatebox{90}{SAM$\downarrow$ (deg)}} & Bicubic & 3.45 & 5.16 & 7.28 & 4.00 & 4.98 & 5.94 & 7.03 & \underline{8.93} & \underline{9.85} & \underline{11.00} & \underline{12.20} & \underline{13.06} \\
 & LIIF & \textcolor{gray}{3.53} & \textcolor{gray}{5.20} & \textcolor{gray}{7.56} & \textcolor{gray}{4.02} & \textcolor{gray}{4.99} & \textcolor{gray}{6.05} & \textcolor{gray}{7.30} & \textcolor{gray}{9.28} & \textcolor{gray}{10.24} & \textcolor{gray}{11.43} & \textcolor{gray}{12.69} & \textcolor{gray}{13.45} \\
 & LTE & 3.28 & 4.93 & \underline{7.21} & 3.77 & 4.74 & 5.76 & \underline{6.95} & \textcolor{gray}{8.97} & \textcolor{gray}{9.89} & \textcolor{gray}{11.11} & \textcolor{gray}{12.37} & \textcolor{gray}{13.28} \\
 & Meta-SR & 3.24 & 4.93 & 7.26 & 3.77 & 4.74 & 5.77 & 7.00 & \textcolor{gray}{9.02} & \textcolor{gray}{9.98} & \textcolor{gray}{11.25} & \textcolor{gray}{12.54} & \textcolor{gray}{13.43} \\
 & CiaoSR & 3.27 & 4.95 & 7.24 & 3.80 & 4.76 & 5.77 & 6.98 & \textcolor{gray}{9.02} & \textcolor{gray}{10.00} & \textcolor{gray}{11.29} & \textcolor{gray}{12.65} & \textcolor{gray}{13.41} \\
 & SRNO & \underline{3.22} & \underline{4.91} & 7.28 & \underline{3.75} & \underline{4.72} & \underline{5.75} & 7.01 & \textcolor{gray}{9.06} & \textcolor{gray}{10.03} & \textcolor{gray}{11.22} & \textcolor{gray}{12.49} & \textcolor{gray}{13.65} \\
 & GaussianSR & \textcolor{gray}{3.45} & 5.05 & \textcolor{gray}{7.38} & 3.92 & 4.86 & 5.89 & \textcolor{gray}{7.12} & \textcolor{gray}{9.09} & \textcolor{gray}{10.03} & \textcolor{gray}{11.24} & \textcolor{gray}{12.52} & \textcolor{gray}{13.40} \\
 & \cellcolor{oursrow}\ours{} & \cellcolor{oursrow}\textbf{3.09} & \cellcolor{oursrow}\textbf{4.70} & \cellcolor{oursrow}\textbf{6.62} & \cellcolor{oursrow}\textbf{3.63} & \cellcolor{oursrow}\textbf{4.55} & \cellcolor{oursrow}\textbf{5.42} & \cellcolor{oursrow}\textbf{6.41} & \cellcolor{oursrow}\textbf{8.37} & \cellcolor{oursrow}\textbf{9.40} & \cellcolor{oursrow}\textbf{10.61} & \cellcolor{oursrow}\textbf{11.99} & \cellcolor{oursrow}\textbf{12.89} \\
\bottomrule
\end{tabular}
\end{adjustbox}
\vspace{-6pt}
\end{table}

\section{Experiments}
\label{sec:exp}

\subsection{Experimental Setup}
\label{sec:exp:setup}

\paragraph{Datasets.} \ours{} is trained on the 900 ARAD training images alone, and the other six datasets, CAVE~\citep{cave}, Harvard~\citep{harvard}, Pavia~U and Pavia~C~\citep{pavia}, Chikusei~\citep{chikusei} and Botswana~\citep{botswana}, serve only for testing. They come from five sensors unseen in training, with 31 to 145 bands spanning the visible to the short-wave infrared.

\paragraph{Baselines.} We compare with six arbitrary-scale methods, the factor-conditioned Meta-SR~\citep{metasr}, the coordinate-query methods LIIF~\citep{liif}, LTE~\citep{lte}, CiaoSR~\citep{ciaosr} and SRNO~\citep{srno}, and the Gaussian-based GaussianSR~\citep{gaussiansr}. They use the ESSAformer encoder~\citep{essaformer} of \ours{} at twice its width or more and are trained on ARAD for 2000 epochs. Their output dimension equals the training band count, so on a new sensor they run on sliding windows of $31$ adjacent bands, with no target-domain training. We also train every baseline on the target scene of Pavia~U and Chikusei, from scratch or through two $1{\times}1$ band adapters on its frozen ARAD body, and test it on tiles that do not overlap the training tiles.

\paragraph{Metrics and protocol.} Every low-resolution input is generated by antialiased bicubic downsampling applied identically to all bands, the shared degradation assumed in Section~\ref{sec:method:prior}. We report PSNR and the spectral angle SAM in degrees. Tables~\ref{tab:main} and~\ref{tab:ssim_sam} use $256{\times}256$ crops of ARAD, Pavia~U and Chikusei at twelve factors, of which $\times12$ to $\times48$ lie outside the training range, and Table~\ref{tab:seven} uses center crops for close-range data and $64{\times}64$ tiles for remote-sensing scenes; values are compared within one table only.

\paragraph{Implementation details.} \ours{} uses the ESSAformer encoder with $C{=}32$, has $0.538$M parameters and is trained for 2000 epochs with Adam, batch size 2, $256{\times}256$ patches and factors drawn uniformly from $[1.5, 8]$, at a learning rate of $2{\times}10^{-4}$ halved at epochs 600, 1200 and 1600. The loss adds a degradation-consistency term, a spectral-angle term and a spectral-curvature term to a Charbonnier reconstruction term.

\vspace{-6pt}
\subsection{Comparison on Unseen Sensors}
\label{sec:exp:main}

\begin{figure}[t]
\centering
\includegraphics[width=\linewidth]{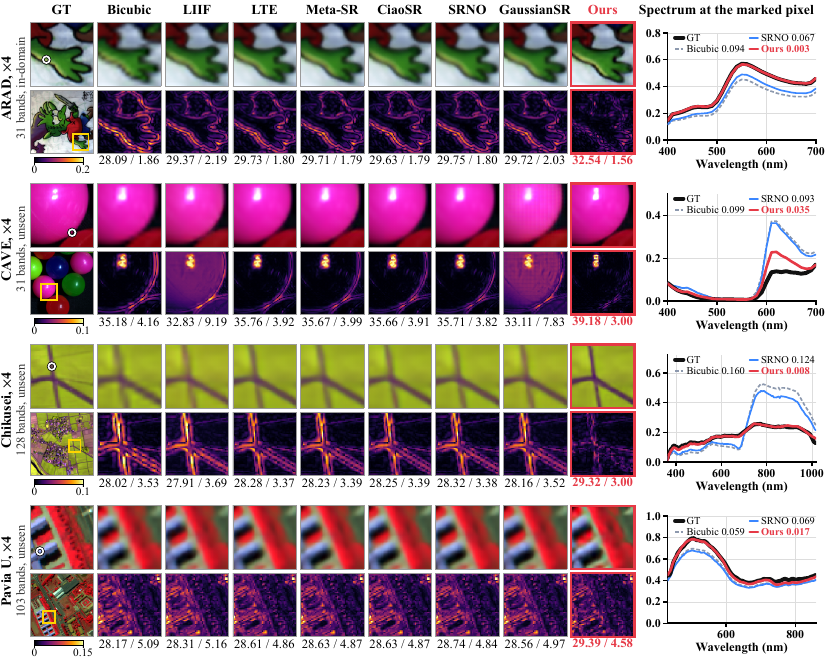}
\vspace{-12pt}
\caption{Visual comparison at $\times$4 on ARAD (in-domain) and on CAVE, Chikusei and Pavia~U (unseen sensors, zero-shot). For each dataset, the top row shows a false-color zoom, Chikusei in the near-infrared composite of bands 70, 100 and 36 and Pavia~U in that of about 810, 650 and 550~nm, and the bottom row the absolute error averaged over bands; the first column gives the full $256{\times}256$ crop with the zoom window. Numbers are PSNR / SAM on the crop. The right column plots the spectrum at the circled pixel with its RMSE in the legend.}
\label{fig:qual}
\vspace{-10pt}
\end{figure}

\paragraph{Quantitative comparison.} As shown in Table~\ref{tab:main}, with every model trained on ARAD alone, \ours{} is the best at all twelve factors on the in-domain ARAD and on both unseen sensors. For example, it leads the strongest baseline SRNO by 1.84~dB on ARAD at $\times4$. On Pavia~U and Chikusei it leads the best baseline at each factor by 0.10 to 1.05~dB and by 0.07 to 1.63~dB, and by 0.55~dB on average over the twelve factors of both sensors. All six baselines fall below bicubic interpolation from $\times36$ on ARAD, $\times24$ on Chikusei and $\times12$ on Pavia~U, and \ours{} never does. In Table~\ref{tab:ssim_sam}, \ours{} has the lowest SAM at all twelve factors on Pavia~U and the highest SSIM at eleven, level with bicubic interpolation at $\times48$. It is also the fastest on all three datasets in the time column, and on the $145$-band Botswana at $\times16$ it runs $36.5\times$ faster than CiaoSR (Figure~\ref{fig:teaser}d).

\paragraph{Seven datasets.} Table~\ref{tab:main} tests two unseen sensors at twelve factors, and Table~\ref{tab:seven} tests one model on many unseen sensors. It runs the same ARAD model on all seven datasets at $\times4$ against the best of the six baselines in each metric, which run on sliding windows of $31$ bands. \ours{} leads on all seven datasets, by 0.32 to 2.07~dB in PSNR and by 0.10 to 0.40$^\circ$ in SAM, and it is better than bicubic interpolation in both metrics everywhere, while the best baseline falls below bicubic interpolation in PSNR on Botswana and in SAM on Harvard and Botswana.

\paragraph{Training on the target sensor.} Training the baselines on the target scene does not close the gap. On the test tiles of Pavia~U and Chikusei, the best baseline trained from scratch or through band adapters stays 0.10 to 1.03~dB below the zero-shot \ours{} at all six factors from $\times3.5$ to $\times12$.

\paragraph{Qualitative comparison.} Figure~\ref{fig:qual} compares the reconstructions at $\times4$ at the level of pixels and spectra. On the in-domain ARAD, the error map of \ours{} is clearly darker along the outlines. On the unseen CAVE, GaussianSR leaves a grid pattern and the other baselines keep visible errors along the outline of the ball, while the error map of \ours{} stays dark. At the circled pixel on the edge of the ball, the baselines mix in the spectrum of the neighboring red ball above 600~nm. On the unseen Chikusei, the error maps of the baselines are almost identical to that of bicubic interpolation, and the thin roads between the paddy fields blur into wide bands, which \ours{} recovers thin and continuous. At the circled road pixel, the baselines mix in the spectra of the neighboring fields, and the spectrum of \ours{} coincides with the ground truth. On Pavia~U, \ours{} renders the edges of the building rows sharper, and at the circled pixel its spectral RMSE is 0.017 against 0.069 for SRNO.

\subsection{Ablation Study}
\label{sec:exp:abl}

\begin{table}[t]
\centering
\begin{minipage}[t]{0.52\linewidth}
\centering
\caption{Ablation at $\times$4; time and memory on Chikusei. Spectral values: the network outputs the value of every band in place of the operators.}
\label{tab:ablation}
\footnotesize\renewcommand{\arraystretch}{1.12}\setlength{\tabcolsep}{2.4pt}
\begin{adjustbox}{max width=\linewidth,center}
\begin{tabular}{@{}l*{6}{c}@{}}
\toprule
\multirow{2}{*}{Variant} & \multicolumn{2}{c}{Pavia~U (103)} & \multicolumn{2}{c}{Chikusei (128)} & \multirow{2}{*}{ms} & \multirow{2}{*}{MiB} \\
\cmidrule(lr){2-3}\cmidrule(lr){4-5}
 & PSNR$\uparrow$ & SAM$\downarrow$ & PSNR$\uparrow$ & SAM$\downarrow$ & & \\
\midrule
Bicubic & 28.13 & 5.16 & 27.60 & 3.95 & -- & -- \\
\midrule
Spectral values & \textcolor{gray}{27.84} & \textcolor{gray}{6.86} & \textcolor{gray}{25.31} & \textcolor{gray}{8.59} & 14.8 & 441 \\
w/o zero-sum & 28.82 & 4.89 & 28.68 & 3.55 & 16.4 & 524 \\
w/o COFR & 28.41 & 5.09 & 27.86 & 3.89 & 14.2 & 344 \\
w/o raw bands & 28.92 & 4.75 & 28.79 & 3.62 & 13.9 & 651 \\
\midrule
\cellcolor{oursrow}\ours{} & \cellcolor{oursrow}\textbf{28.94} & \cellcolor{oursrow}\textbf{4.70} & \cellcolor{oursrow}\textbf{28.95} & \cellcolor{oursrow}\textbf{3.38} & \cellcolor{oursrow}32.2 & \cellcolor{oursrow}716 \\
\bottomrule
\end{tabular}
\end{adjustbox}
\end{minipage}\hfill
\begin{minipage}[t]{0.455\linewidth}
\centering
\caption{Zero-shot comparison at $\times$4 with the best of the six baselines in each metric; ARAD is in-domain.}
\label{tab:seven}
\footnotesize\renewcommand{\arraystretch}{1.192}\setlength{\tabcolsep}{2.4pt}
\begin{adjustbox}{max width=\linewidth,center}
\begin{tabular}{@{}l*{3}{c}@{}}
\toprule
\multirow{2}{*}{Dataset} & \multicolumn{3}{c}{PSNR$\uparrow$ / SAM$\downarrow$} \\
\cmidrule(lr){2-4}
 & Bicubic & Best baseline & \ours{} \\
\midrule
ARAD (31) & 37.50\,/\,1.74 & 38.49\,/\,1.63 & \cellcolor{oursrow}\textbf{40.33}\,/\,\textbf{1.38} \\
\midrule
CAVE (31) & 34.88\,/\,3.12 & 35.59\,/\,2.93 & \cellcolor{oursrow}\textbf{37.66}\,/\,\textbf{2.53} \\
Harvard (31) & 43.09\,/\,2.38 & 43.53\,/\,\textcolor{gray}{2.41} & \cellcolor{oursrow}\textbf{44.85}\,/\,\textbf{2.28} \\
Pavia~U (103) & 29.04\,/\,5.38 & 29.40\,/\,5.17 & \cellcolor{oursrow}\textbf{29.98}\,/\,\textbf{4.91} \\
Pavia~C (102) & 30.05\,/\,6.39 & 30.43\,/\,6.19 & \cellcolor{oursrow}\textbf{31.11}\,/\,\textbf{5.97} \\
Chikusei (128) & 29.10\,/\,11.09 & 29.73\,/\,10.93 & \cellcolor{oursrow}\textbf{30.92}\,/\,\textbf{10.65} \\
Botswana (145) & 53.07\,/\,2.38 & \textcolor{gray}{52.94}\,/\,\textcolor{gray}{2.44} & \cellcolor{oursrow}\textbf{53.26}\,/\,\textbf{2.34} \\
\bottomrule
\end{tabular}
\end{adjustbox}
\end{minipage}
\vspace{-6pt}
\end{table}

\paragraph{Ablation.} Table~\ref{tab:ablation} changes the output of the network and removes the key designs of Section~\ref{sec:method} one at a time on the two unseen sensors. Outputting the spectral values outright, with no observed pixels combined, falls below bicubic interpolation on both sensors, by 2.3~dB on Chikusei. Without COFR the operators act once on the low-resolution grid and the result is upsampled bicubically, which costs 0.53~dB on Pavia~U and 1.09~dB on Chikusei. Dropping the zero-sum constraint costs 0.12 and 0.27~dB, and applying the operators to $31$ resampled bands costs 0.02 and 0.16~dB.

\paragraph{Band sharing.} Section~\ref{sec:method:prior} assumes that one operator serves all bands. A zero-sum $5{\times}5$ operator fitted by least squares, with no network, on four bands of every $16{\times}16$ window is at least as accurate on the other bands as per-band operators, on held-out pixels of all seven datasets at $\times2$ and $\times4$, and one fitted on a single band loses at most 0.01~dB on six of them and up to 0.29~dB on CAVE. Eq.~(\ref{eq:equivariance}) predicts that redefining the bands passes through the reconstruction, and keeping every other band, averaging neighboring pairs or applying a per-band gain or offset before \ours{} matches applying it afterwards within 0.05~dB on all seven datasets.

\section{Conclusion}
\label{sec:conclusion}

This paper shows that success in cross-sensor HSR depends on what the network outputs. With the backbone and training shared, a network that outputs spectral values is less accurate than one that predicts band-shared spatial operators on every dataset, even when it serves any band count (Section~\ref{sec:exp:abl}), and adapting its interface or training it on the target sensor still leaves it less accurate than a model that has never seen that sensor (Section~\ref{sec:exp:main}). \ours{} only decides how the observed pixels are combined, transfers from the $31$ bands of ARAD to sensors with up to $145$ bands without target data (Table~\ref{tab:seven}), and keeps its lead at factors far beyond its training range (Table~\ref{tab:main}). The spectrum of a new sensor is already in its observation, and what is worth learning is how its pixels are combined in space. One operator shared by all bands assumes that they share their spatial blur and registration, and bands that differ in either call for operators that vary across bands. The low-resolution inputs in this paper are all synthesized, and real or noisy observations remain to be tested.

\section*{AI Use Statement}
We have not used generative AI tools for any task that requires disclosure under the ICLR 2027 AI Policy for Authors: no such tool was used to develop the method or its conceptual framework, to formulate hypotheses or claims, to design the experiments, to implement the method, to generate or process data, to translate, or to interpret the results.
We used generative AI tools only to edit and rephrase the text of this paper for readability.
We have reviewed all AI-assisted edits, and we take responsibility for the final content of this work, including text, claims or artifacts produced with the aid of generative AI.

\section*{Ethics statement}

This work uses only the public ARAD, CAVE, Harvard, Pavia University, Pavia Centre, Chikusei and Botswana hyperspectral datasets under their original terms of use and involves no new data collection or human subjects. We are not aware of ethical concerns specific to this work.

\section*{Reproducibility statement}
All experiments use the seven public datasets listed in Appendix~\ref{app:datasets}, with the crops, the bicubic degradation and the metrics described in Section~\ref{sec:exp:setup}. The same appendix specifies the parameterization of the primitives, the rendering width, the losses and their weights, the operator window, the initialization and the random seed of \ours{}, and Section~\ref{sec:exp:setup} gives its optimizer, learning-rate schedule and number of epochs. Appendix~\ref{app:routes} specifies how every baseline is used on a new sensor, directly, trained from scratch or through band adapters, and Appendix~\ref{sec:app:efficiency} gives the protocol of the time measurements. The code will be publicly released.

\bibliography{iclr2027_conference}
\bibliographystyle{iclr2027_conference}

\clearpage
\raggedbottom
\appendix
\renewcommand{\topfraction}{0.9}\renewcommand{\bottomfraction}{0.8}\renewcommand{\textfraction}{0.1}\renewcommand{\floatpagefraction}{0.8}
\setcounter{topnumber}{3}\setcounter{bottomnumber}{2}\setcounter{totalnumber}{4}
\makeatletter\setlength{\@fptop}{0pt}\setlength{\@fpsep}{\floatsep}\setlength{\@fpbot}{0pt plus 1fil}\makeatother
\begin{center}
{\Large\sc Appendix}
\end{center}
\section{Datasets and Implementation Details}
\label{app:datasets}
\label{app:impl}

\paragraph{Datasets.} Table~\ref{tab:datasets} lists the sensor, platform, band count, spectral range and test size of the seven datasets. Tables~\ref{tab:main} and~\ref{tab:ssim_sam} use the center $256{\times}256$ crop of each of the 50 ARAD test images, six crops on a $3{\times}2$ grid over the $610{\times}340$ Pavia~U scene and the eight Chikusei test tiles, with a low-resolution side of $\mathrm{round}(256/s)$ at factor $s$. Every low-resolution input is obtained by antialiased bicubic downsampling, and every model receives the effective factor, the ratio of the high- to the low-resolution side. The seven-dataset evaluation uses center $256{\times}256$ crops for close-range data and $64{\times}64$ tiles for remote-sensing scenes.

\begin{table}[!htbp]
\centering
\caption{Datasets. \ours{} is trained on ARAD alone; every other dataset is used for testing only, with no target-domain training. Test counts are images for close-range data and non-overlapping $64{\times}64$ tiles for remote-sensing scenes. Spectral ranges are those of the sensors; the model receives normalized band positions and no wavelength. Botswana keeps 145 of the 242 Hyperion bands after the removal of water-absorption and noisy bands, so its bands are not evenly spaced in wavelength.}
\label{tab:datasets}
\footnotesize\renewcommand{\arraystretch}{1.12}\setlength{\tabcolsep}{5pt}
\begin{adjustbox}{max width=\linewidth,center}
\begin{tabular}{@{}l l l c c c@{}}
\toprule
Dataset & Sensor & Platform & Bands & Range (nm) & Test \\
\midrule
ARAD~\citep{arad2022ntire} & Specim IQ & ground & 31 & 400--700 & 50 \\
CAVE~\citep{cave} & CCD with tunable filter & laboratory & 31 & 400--700 & 31 \\
Harvard~\citep{harvard} & Nuance FX & ground & 31 & 420--720 & 16 \\
Pavia~U~\citep{pavia} & ROSIS-03 & airborne & 103 & 430--860 & 10 / 45 \\
Pavia~C~\citep{pavia} & ROSIS-03 & airborne & 102 & 430--860 & 187 \\
Chikusei~\citep{chikusei} & Headwall Hyperspec-VNIR-C & airborne & 128 & 363--1018 & 311 \\
Botswana~\citep{botswana} & Hyperion (EO-1) & spaceborne & 145 & 400--2500 & 92 \\
\bottomrule
\end{tabular}
\end{adjustbox}
\end{table}

\paragraph{Forward pass.} Algorithm~\ref{alg:omnihsr} lists the forward pass of \ours{} at a factor $s$, and the paragraphs below give the details of each step.

\begin{algorithm}[!htbp]
\caption{Forward pass of \ours{} at a factor $s$.}
\label{alg:omnihsr}
\SetCommentSty{textnormal}\SetArgSty{textnormal}
\KwIn{$X$ : $(h, w, B)$, factor $s$}
\KwOut{$\hat{Y}$ : $(sh, sw, B)$}
$Z \leftarrow C_B X$ \tcp*[l]{$[h \times w \times M]$, Eq.~(\ref{eq:canon})}
$F_s \leftarrow \gamma(s) \odot E_\theta(Z) + \beta(s)$\;
$\{(\mu_n, \Sigma_n, \alpha_n, \tilde{\mathbf{k}}_n)\}_{n=1}^{hw} \leftarrow \mathrm{Head}(F_s)$ \tcp*[l]{one primitive per pixel $q_n$}
$\mathbf{k}_n \leftarrow \tilde{\mathbf{k}}_n - \mathrm{mean}(\tilde{\mathbf{k}}_n)$ \tcp*[l]{$[hw \times 25]$, Eq.~(\ref{eq:zerosum})}
\For{$n \in [1, hw]$}{
  $\mathbf{r}_n \leftarrow \sum_{q \in \mathcal{N}_n} k_{n,q}\, \mathbf{x}(q)$ \tcp*[l]{$[B]$, Eq.~(\ref{eq:anchor})}
}
$\hat{Y} \leftarrow U_s X$ \tcp*[l]{bicubic, $[sh \times sw \times B]$}
\For{each target position $p$, spaced $1/s$ apart}{
  $\omega_n(p) \leftarrow \alpha_n G_n(p) \,/\, \bigl(\sum_j \alpha_j G_j(p) + \epsilon\bigr)$ \tcp*[l]{Eqs.~(\ref{eq:render}), (\ref{eq:sigma_eff})}
  $\hat{Y}(p) \leftarrow \hat{Y}(p) + \sum_n \omega_n(p)\, \mathbf{r}_n$ \tcp*[l]{all $B$ bands}
}
\end{algorithm}

\paragraph{Primitives.} The center of primitive $n$ is its pixel center plus an offset of $0.5\tanh(\cdot)$ low-resolution pixels. The covariance is $\Sigma_n = R(\theta_n)\,\mathrm{diag}(\sigma_{n,1}^2, \sigma_{n,2}^2)\,R(\theta_n)^{\top}$ with the rotation angle $\theta_n = \pi\tanh(\cdot)$ and the widths $\sigma_{n,i} = \mathrm{softplus}(\cdot) + 0.05$ clipped to $[0.05, 2.5]$ low-resolution pixels, and the opacity is $\alpha_n = \mathrm{sigmoid}(\cdot)$.

\paragraph{Rendering width.} The width that each Gaussian uses at rendering carries an analytic anti-aliasing floor that shrinks with the factor and holds no learnable parameter,
\begin{equation}
\label{eq:sigma_eff}
\sigma_{\mathrm{eff}} = \sqrt{\, \sigma^2 + (a/s)^2 \,},
\end{equation}
where $\sigma$ is the predicted width and $a = 0.5$. Each primitive is rendered in a square window of radius $\lceil 3\sigma_{\max} s \rceil$ high-resolution pixels, where $\sigma_{\max}$ is the largest $\sigma_{\mathrm{eff}}$ in the image, and $\epsilon = 10^{-6}$ in Eq.~(\ref{eq:render}).

\paragraph{Training.} The reconstruction term is the Charbonnier loss $\sqrt{(\hat{Y}-Y)^2 + 10^{-6}}$ averaged over pixels and bands. The degradation-consistency term downsamples $\hat{Y}$ to the input size with antialiased bicubic interpolation and applies the same loss against $X$. The spectral-angle term averages the angle between the predicted and the true spectrum over pixels, with the cosine clipped to $[-1+10^{-6}, 1-10^{-6}]$, and the spectral-curvature term is the mean absolute difference between the second differences of $\hat{Y}$ and $Y$ along the band axis. The degradation-consistency, spectral-angle and spectral-curvature terms are weighted $0.1$, $0.1$ and $0.05$, and training runs on one RTX~4090D with seed 2026.

\paragraph{Operator window and initialization.} The $5\times5$ window uses replicate padding at the image border, and Appendix~\ref{app:window} compares other window sizes. The output layer that predicts the operators has zero-initialized weights and bias, so every operator starts at zero and the initial reconstruction operator $W_s$ is bicubic interpolation.

\FloatBarrier
\section{Comparison with Baselines}
\label{app:results}

\subsection{The Three Existing Routes}
\label{app:routes}

\begin{figure}[!htbp]
\centering
\includegraphics[width=0.55\linewidth]{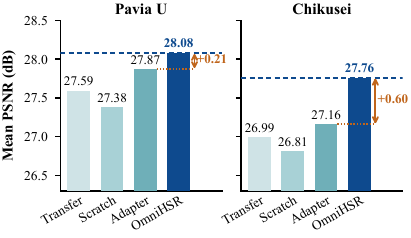}
\caption{Zero-shot \ours{} against the best of the six baselines under each way of reusing them on a new sensor, mean PSNR (dB) over seven factors from $\times$3.5 to $\times$16 on test tiles of Pavia~U and Chikusei. Orange: lead of \ours{} over the best of the three ways.}
\label{fig:routes}
\end{figure}

Figure~\ref{fig:routes} and Table~\ref{tab:routes_full} compare the three ways to use a baseline on a new sensor. They keep the protocol of the target-trained baselines, the $64{\times}64$ test tiles of Pavia~U and Chikusei that do not overlap their training tiles, with a low-resolution side of $\mathrm{round}(64/s)$. Direct transfer runs the ARAD model on sliding windows of 31 adjacent bands and averages the overlaps, with no target-domain training. Training from scratch follows each method's original recipe on the target scene for 2000 epochs. Adapter tuning freezes the ARAD body, attaches one $1{\times}1$ convolution before it and one after it to convert the band count, trains only these two layers from an exact linear band resampling, and runs until the learning rate has annealed.

Figure~\ref{fig:routes} averages each way over the seven factors from $\times3.5$ to $\times16$, taking the best of the six baselines at each factor. Adapter tuning is the strongest of the three ways, and the zero-shot \ours{} stays above it by 0.21~dB on Pavia~U and by 0.60~dB on Chikusei. It also has the highest PSNR of all eighteen configurations at every factor, by 0.03 to 1.03~dB. Direct transfer lifts the stronger methods above bicubic interpolation at small factors, and all six methods fall below it at $\times12$ on Pavia~U and at $\times16$ on both scenes. Training from scratch leaves all six methods below direct transfer of the same model at $\times12$ on Chikusei, most for SRNO at 22.87 against 24.60~dB. Adapter tuning lifts every method, and the size of the lift is almost entirely set by the starting point, with a correlation of $-0.93$ between the two at $\times4$ on Pavia~U and $-0.99$ on Chikusei. After adapter tuning, LIIF remains below bicubic interpolation at all seven factors on Chikusei, and only Meta-SR among the six methods is above it at $\times12$ on Pavia~U and at $\times16$ on both scenes.

The main model can also be run as a $31$-band model on sliding windows: under the protocol of Table~\ref{tab:main} at $\times4$, PSNR rises from 28.94 to 29.12~dB on Pavia~U and from 28.95 to 29.13~dB on Chikusei, at the cost of $7$ to $10$ forward passes, 3.6 to 3.8 times the latency, and the requirement of at least $31$ bands.

\begin{table}[!htbp]
\centering
\caption{The three ways to reuse a baseline on a new sensor at seven factors from $\times$3.5 to $\times$16 (PSNR$\uparrow$, dB), the data behind Figure~\ref{fig:routes}. Transfer: the ARAD model run on sliding windows of $31$ bands, no target training. Scratch: trained on the target scene for 2000 epochs. Adapter: ARAD body frozen, two $1{\times}1$ band adapters tuned on the target scene. Gray: below bicubic interpolation.}
\label{tab:routes_full}
\footnotesize\renewcommand{\arraystretch}{1.12}\setlength{\tabcolsep}{2.8pt}
\begin{adjustbox}{max width=\linewidth,center}
\begin{tabular}{@{}l l *{14}{c}@{}}
\toprule
\multirow{2}{*}{Method} & \multirow{2}{*}{Route} & \multicolumn{7}{c}{Pavia~U ($103$ bands)} & \multicolumn{7}{c}{Chikusei ($128$ bands)} \\
\cmidrule(lr){3-9}\cmidrule(lr){10-16}
 & & $\times$3.5 & $\times$4 & $\times$5.5 & $\times$8 & $\times$9.5 & $\times$12 & $\times$16 & $\times$3.5 & $\times$4 & $\times$5.5 & $\times$8 & $\times$9.5 & $\times$12 & $\times$16 \\
\midrule
Bicubic & & 30.62 & 29.94 & 28.52 & 26.89 & 26.38 & 25.48 & 24.96 & 29.98 & 29.21 & 27.67 & 26.05 & 25.59 & 24.60 & 24.04 \\
\midrule
\multirow{3}{*}{LIIF} & Transfer & \textcolor{gray}{30.23} & \textcolor{gray}{29.53} & \textcolor{gray}{28.10} & \textcolor{gray}{26.50} & \textcolor{gray}{26.05} & \textcolor{gray}{25.19} & \textcolor{gray}{24.70} & \textcolor{gray}{29.70} & \textcolor{gray}{28.94} & \textcolor{gray}{27.42} & \textcolor{gray}{25.81} & \textcolor{gray}{25.34} & \textcolor{gray}{24.34} & \textcolor{gray}{23.77} \\
 & Scratch & \textcolor{gray}{29.69} & \textcolor{gray}{29.09} & \textcolor{gray}{27.83} & \textcolor{gray}{26.28} & \textcolor{gray}{25.88} & \textcolor{gray}{25.03} & \textcolor{gray}{24.58} & 30.50 & 29.53 & \textcolor{gray}{27.59} & \textcolor{gray}{25.73} & \textcolor{gray}{25.21} & \textcolor{gray}{23.98} & \textcolor{gray}{23.36} \\
 & Adapter & 30.67 & 29.99 & 28.54 & 26.90 & \textcolor{gray}{26.37} & \textcolor{gray}{25.36} & \textcolor{gray}{24.83} & \textcolor{gray}{29.89} & \textcolor{gray}{29.14} & \textcolor{gray}{27.61} & \textcolor{gray}{25.96} & \textcolor{gray}{25.47} & \textcolor{gray}{24.44} & \textcolor{gray}{23.85} \\
\midrule
\multirow{3}{*}{LTE} & Transfer & 30.86 & 30.15 & 28.63 & \textcolor{gray}{26.82} & \textcolor{gray}{26.34} & \textcolor{gray}{25.34} & \textcolor{gray}{24.80} & 30.53 & 29.70 & 28.03 & 26.23 & 25.72 & 24.62 & \textcolor{gray}{23.99} \\
 & Scratch & \textcolor{gray}{30.39} & \textcolor{gray}{29.71} & \textcolor{gray}{28.23} & \textcolor{gray}{26.47} & \textcolor{gray}{26.02} & \textcolor{gray}{25.16} & \textcolor{gray}{24.75} & 30.60 & 29.60 & \textcolor{gray}{27.64} & \textcolor{gray}{25.63} & \textcolor{gray}{25.09} & \textcolor{gray}{23.78} & \textcolor{gray}{23.07} \\
 & Adapter & 31.14 & 30.42 & 28.89 & 27.09 & 26.58 & \textcolor{gray}{25.46} & \textcolor{gray}{24.91} & 30.72 & 29.91 & 28.24 & 26.41 & 25.89 & 24.72 & \textcolor{gray}{24.02} \\
\midrule
\multirow{3}{*}{Meta-SR} & Transfer & 30.83 & 30.10 & 28.57 & \textcolor{gray}{26.80} & \textcolor{gray}{26.32} & \textcolor{gray}{25.33} & \textcolor{gray}{24.80} & 30.43 & 29.59 & 27.92 & 26.14 & 25.63 & \textcolor{gray}{24.57} & \textcolor{gray}{23.97} \\
 & Scratch & \textcolor{gray}{30.28} & \textcolor{gray}{29.59} & \textcolor{gray}{28.20} & \textcolor{gray}{26.48} & \textcolor{gray}{26.02} & \textcolor{gray}{25.15} & \textcolor{gray}{24.72} & 30.78 & 29.88 & 27.99 & \textcolor{gray}{26.01} & \textcolor{gray}{25.44} & \textcolor{gray}{24.13} & \textcolor{gray}{23.44} \\
 & Adapter & 31.17 & 30.49 & 28.96 & 27.15 & 26.61 & 25.52 & 24.98 & 30.67 & 29.85 & 28.18 & 26.38 & 25.87 & 24.76 & 24.11 \\
\midrule
\multirow{3}{*}{CiaoSR} & Transfer & 30.80 & 30.07 & \textcolor{gray}{28.48} & \textcolor{gray}{26.63} & \textcolor{gray}{26.11} & \textcolor{gray}{25.14} & \textcolor{gray}{24.77} & 30.44 & 29.60 & 27.90 & 26.08 & \textcolor{gray}{25.55} & \textcolor{gray}{24.46} & \textcolor{gray}{23.91} \\
 & Scratch & \textcolor{gray}{30.12} & \textcolor{gray}{29.50} & \textcolor{gray}{28.16} & \textcolor{gray}{26.58} & \textcolor{gray}{26.15} & \textcolor{gray}{25.31} & \textcolor{gray}{24.82} & 30.56 & 29.58 & \textcolor{gray}{27.44} & \textcolor{gray}{25.22} & \textcolor{gray}{24.75} & \textcolor{gray}{23.62} & \textcolor{gray}{23.02} \\
 & Adapter & 31.07 & 30.38 & 28.78 & 26.98 & 26.45 & \textcolor{gray}{25.36} & \textcolor{gray}{24.83} & 30.66 & 29.87 & 28.19 & 26.35 & 25.81 & 24.66 & \textcolor{gray}{24.01} \\
\midrule
\multirow{3}{*}{SRNO} & Transfer & 30.94 & 30.21 & 28.65 & \textcolor{gray}{26.81} & \textcolor{gray}{26.28} & \textcolor{gray}{24.98} & \textcolor{gray}{24.05} & 30.56 & 29.73 & 28.04 & 26.24 & 25.72 & 24.60 & \textcolor{gray}{23.91} \\
 & Scratch & 30.66 & \textcolor{gray}{29.91} & \textcolor{gray}{28.23} & \textcolor{gray}{26.31} & \textcolor{gray}{25.85} & \textcolor{gray}{24.92} & \textcolor{gray}{24.48} & 30.56 & 29.61 & \textcolor{gray}{27.57} & \textcolor{gray}{25.36} & \textcolor{gray}{24.68} & \textcolor{gray}{22.87} & \textcolor{gray}{21.80} \\
 & Adapter & 31.27 & 30.56 & 28.97 & 27.12 & 26.59 & \textcolor{gray}{25.39} & \textcolor{gray}{24.87} & 30.78 & 29.95 & 28.24 & 26.41 & 25.88 & 24.72 & \textcolor{gray}{23.93} \\
\midrule
\multirow{3}{*}{GaussianSR} & Transfer & 30.78 & 30.08 & \textcolor{gray}{28.52} & \textcolor{gray}{26.75} & \textcolor{gray}{26.26} & \textcolor{gray}{25.30} & \textcolor{gray}{24.78} & 30.21 & 29.40 & 27.70 & \textcolor{gray}{25.93} & \textcolor{gray}{25.41} & \textcolor{gray}{24.31} & \textcolor{gray}{23.69} \\
 & Scratch & \textcolor{gray}{29.72} & \textcolor{gray}{29.16} & \textcolor{gray}{27.78} & \textcolor{gray}{26.15} & \textcolor{gray}{25.70} & \textcolor{gray}{24.86} & \textcolor{gray}{24.48} & 30.52 & 29.55 & \textcolor{gray}{27.54} & \textcolor{gray}{25.51} & \textcolor{gray}{24.93} & \textcolor{gray}{23.69} & \textcolor{gray}{23.07} \\
 & Adapter & 31.04 & 30.36 & 28.81 & 27.00 & 26.45 & \textcolor{gray}{25.42} & \textcolor{gray}{24.86} & 30.38 & 29.61 & 27.97 & 26.20 & 25.68 & \textcolor{gray}{24.58} & \textcolor{gray}{23.94} \\
\midrule
\cellcolor{oursrow}\ours{} & \cellcolor{oursrow}zero-shot & \cellcolor{oursrow}\textbf{31.53} & \cellcolor{oursrow}\textbf{30.83} & \cellcolor{oursrow}\textbf{29.19} & \cellcolor{oursrow}\textbf{27.41} & \cellcolor{oursrow}\textbf{26.96} & \cellcolor{oursrow}\textbf{25.62} & \cellcolor{oursrow}\textbf{25.01} & \cellcolor{oursrow}\textbf{31.81} & \cellcolor{oursrow}\textbf{30.92} & \cellcolor{oursrow}\textbf{29.03} & \cellcolor{oursrow}\textbf{26.92} & \cellcolor{oursrow}\textbf{26.33} & \cellcolor{oursrow}\textbf{25.01} & \cellcolor{oursrow}\textbf{24.30} \\
\bottomrule
\end{tabular}
\end{adjustbox}
\end{table}

\FloatBarrier
\subsection{All Baselines on Seven Datasets}
\label{app:breadth}

Table~\ref{tab:breadth_full} is the full version of Table~\ref{tab:seven}, with every baseline at $\times2$, $\times4$ and $\times8$. \ours{} is ahead of every baseline in PSNR and SAM on every dataset at every factor. GaussianSR falls below bicubic interpolation in PSNR on all six unseen datasets at $\times8$ and on CAVE, Harvard and Botswana at all three factors, and LIIF on all six unseen datasets at $\times4$ and $\times8$. On the $145$-band Botswana, which extends to the short-wave infrared, every baseline falls below bicubic interpolation in PSNR at $\times4$ and $\times8$, and on Harvard and Botswana every baseline has a higher SAM than bicubic interpolation at all three factors. Paired over the test images and tiles, the PSNR gain of \ours{} over bicubic interpolation has a 95\% bootstrap interval above zero on every dataset at every factor, and at $\times4$ \ours{} is above SRNO on every image and tile of all seven datasets.

\begin{table}[!htbp]
\centering
\caption{One ARAD-trained model on seven datasets at $\times$2, $\times$4 and $\times$8 (PSNR$\uparrow$ / SAM$\downarrow$); the full version of Table~\ref{tab:seven} with every baseline. Baselines run on sliding windows of $31$ bands with no target training; ARAD is in-domain. Remote-sensing scenes are scored on $64{\times}64$ test tiles, not on the $256{\times}256$ crops of Table~\ref{tab:main}. Gray: worse than bicubic interpolation. \textbf{Bold}: best.}
\label{tab:breadth_full}
\footnotesize\renewcommand{\arraystretch}{1.12}\setlength{\tabcolsep}{2.4pt}
\begin{adjustbox}{max width=\linewidth,center}
\begin{tabular}{@{}l l *{7}{c}@{}}
\toprule
 & \multirow{2}{*}{Method} & ARAD & CAVE & Harvard & Pavia~U & Pavia~C & Chikusei & Botswana \\
 &  & (31) & (31) & (31) & (103) & (102) & (128) & (145) \\
\midrule
\multirow{8}{*}{$\times$2} & Bicubic & 44.04\,/\,0.91 & 40.83\,/\,1.95 & 47.47\,/\,1.90 & 33.49\,/\,3.56 & 34.60\,/\,4.60 & 34.62\,/\,9.46 & 56.31\,/\,1.64 \\
 & LIIF & 44.48\,/\,\textcolor{gray}{1.22} & \textcolor{gray}{37.03}\,/\,\textcolor{gray}{5.56} & \textcolor{gray}{41.33}\,/\,\textcolor{gray}{7.16} & \textcolor{gray}{33.39}\,/\,\textcolor{gray}{3.74} & 34.61\,/\,\textcolor{gray}{4.83} & \textcolor{gray}{34.41}\,/\,\textcolor{gray}{9.73} & \textcolor{gray}{50.34}\,/\,\textcolor{gray}{5.12} \\
 & LTE & 45.33\,/\,0.90 & 41.44\,/\,1.90 & 47.76\,/\,\textcolor{gray}{1.97} & 34.13\,/\,3.40 & 35.27\,/\,4.44 & 35.64\,/\,9.34 & \textcolor{gray}{56.22}\,/\,\textcolor{gray}{1.68} \\
 & Meta-SR & 45.42\,/\,0.87 & 41.51\,/\,1.87 & 47.71\,/\,\textcolor{gray}{1.96} & 34.24\,/\,3.36 & 35.38\,/\,4.40 & 35.72\,/\,9.30 & 56.36\,/\,\textcolor{gray}{1.65} \\
 & CiaoSR & 45.39\,/\,0.88 & 41.27\,/\,1.91 & 47.78\,/\,\textcolor{gray}{1.97} & 34.14\,/\,3.39 & 35.29\,/\,4.44 & 35.68\,/\,9.34 & \textcolor{gray}{56.17}\,/\,\textcolor{gray}{1.70} \\
 & SRNO & 45.58\,/\,0.87 & 41.49\,/\,1.89 & 47.88\,/\,\textcolor{gray}{1.96} & 34.34\,/\,3.34 & 35.48\,/\,4.38 & 35.81\,/\,9.32 & \textcolor{gray}{56.22}\,/\,\textcolor{gray}{1.69} \\
 & GaussianSR & 44.71\,/\,\textcolor{gray}{1.16} & \textcolor{gray}{37.64}\,/\,\textcolor{gray}{5.00} & \textcolor{gray}{41.04}\,/\,\textcolor{gray}{6.96} & 34.02\,/\,\textcolor{gray}{3.58} & 35.16\,/\,\textcolor{gray}{4.69} & 35.16\,/\,\textcolor{gray}{9.57} & \textcolor{gray}{50.78}\,/\,\textcolor{gray}{4.92} \\
 & \cellcolor{oursrow}\ours{} & \cellcolor{oursrow}\textbf{48.41}\,/\,\textbf{0.68} & \cellcolor{oursrow}\textbf{44.27}\,/\,\textbf{1.59} & \cellcolor{oursrow}\textbf{49.02}\,/\,\textbf{1.81} & \cellcolor{oursrow}\textbf{35.39}\,/\,\textbf{3.19} & \cellcolor{oursrow}\textbf{36.67}\,/\,\textbf{4.20} & \cellcolor{oursrow}\textbf{37.49}\,/\,\textbf{9.13} & \cellcolor{oursrow}\textbf{57.03}\,/\,\textbf{1.55} \\
\midrule
\multirow{8}{*}{$\times$4} & Bicubic & 37.50\,/\,1.74 & 34.88\,/\,3.12 & 43.09\,/\,2.38 & 29.04\,/\,5.38 & 30.05\,/\,6.39 & 29.10\,/\,11.09 & 53.07\,/\,2.38 \\
 & LIIF & 37.96\,/\,\textcolor{gray}{1.94} & \textcolor{gray}{33.45}\,/\,\textcolor{gray}{6.29} & \textcolor{gray}{39.04}\,/\,\textcolor{gray}{7.39} & \textcolor{gray}{28.81}\,/\,\textcolor{gray}{5.62} & \textcolor{gray}{29.84}\,/\,\textcolor{gray}{6.62} & \textcolor{gray}{28.94}\,/\,\textcolor{gray}{11.32} & \textcolor{gray}{48.58}\,/\,\textcolor{gray}{5.57} \\
 & LTE & 38.38\,/\,1.64 & 35.52\,/\,2.93 & 43.39\,/\,\textcolor{gray}{2.42} & 29.35\,/\,5.17 & 30.41\,/\,6.19 & 29.70\,/\,10.93 & \textcolor{gray}{52.94}\,/\,\textcolor{gray}{2.44} \\
 & Meta-SR & 38.33\,/\,1.66 & 35.41\,/\,2.97 & 43.23\,/\,\textcolor{gray}{2.41} & 29.30\,/\,5.19 & 30.34\,/\,6.21 & 29.59\,/\,10.96 & \textcolor{gray}{52.86}\,/\,\textcolor{gray}{2.45} \\
 & CiaoSR & 38.39\,/\,1.64 & 35.54\,/\,2.94 & 43.51\,/\,\textcolor{gray}{2.42} & 29.27\,/\,5.21 & 30.33\,/\,6.23 & 29.59\,/\,10.97 & \textcolor{gray}{52.85}\,/\,\textcolor{gray}{2.45} \\
 & SRNO & 38.49\,/\,1.63 & 35.59\,/\,2.93 & 43.53\,/\,\textcolor{gray}{2.41} & 29.40\,/\,5.20 & 30.43\,/\,6.24 & 29.73\,/\,10.94 & \textcolor{gray}{52.81}\,/\,\textcolor{gray}{2.46} \\
 & GaussianSR & 38.23\,/\,\textcolor{gray}{1.81} & \textcolor{gray}{33.79}\,/\,\textcolor{gray}{5.48} & \textcolor{gray}{38.65}\,/\,\textcolor{gray}{7.14} & 29.28\,/\,5.30 & 30.33\,/\,6.31 & 29.39\,/\,\textcolor{gray}{11.14} & \textcolor{gray}{49.23}\,/\,\textcolor{gray}{5.33} \\
 & \cellcolor{oursrow}\ours{} & \cellcolor{oursrow}\textbf{40.33}\,/\,\textbf{1.38} & \cellcolor{oursrow}\textbf{37.66}\,/\,\textbf{2.53} & \cellcolor{oursrow}\textbf{44.85}\,/\,\textbf{2.28} & \cellcolor{oursrow}\textbf{29.98}\,/\,\textbf{4.91} & \cellcolor{oursrow}\textbf{31.11}\,/\,\textbf{5.97} & \cellcolor{oursrow}\textbf{30.92}\,/\,\textbf{10.65} & \cellcolor{oursrow}\textbf{53.26}\,/\,\textbf{2.34} \\
\midrule
\multirow{8}{*}{$\times$8} & Bicubic & 33.37\,/\,2.72 & 30.86\,/\,4.71 & 39.56\,/\,2.76 & 26.12\,/\,7.63 & 27.12\,/\,8.18 & 26.00\,/\,12.73 & 51.22\,/\,3.00 \\
 & LIIF & 33.56\,/\,\textcolor{gray}{2.89} & \textcolor{gray}{30.16}\,/\,\textcolor{gray}{7.58} & \textcolor{gray}{36.49}\,/\,\textcolor{gray}{7.65} & \textcolor{gray}{25.77}\,/\,\textcolor{gray}{8.21} & \textcolor{gray}{26.77}\,/\,\textcolor{gray}{8.61} & \textcolor{gray}{25.80}\,/\,\textcolor{gray}{13.04} & \textcolor{gray}{47.14}\,/\,\textcolor{gray}{6.01} \\
 & LTE & 33.87\,/\,2.58 & 31.37\,/\,4.40 & 39.88\,/\,\textcolor{gray}{2.78} & 26.14\,/\,\textcolor{gray}{7.74} & 27.18\,/\,8.14 & 26.23\,/\,12.66 & \textcolor{gray}{51.08}\,/\,\textcolor{gray}{3.06} \\
 & Meta-SR & 33.81\,/\,2.63 & 31.23\,/\,4.47 & 39.66\,/\,\textcolor{gray}{2.79} & \textcolor{gray}{26.09}\,/\,\textcolor{gray}{7.76} & \textcolor{gray}{27.10}\,/\,8.17 & 26.14\,/\,12.70 & \textcolor{gray}{50.94}\,/\,\textcolor{gray}{3.09} \\
 & CiaoSR & 33.86\,/\,2.57 & 31.39\,/\,4.41 & 39.94\,/\,\textcolor{gray}{2.78} & \textcolor{gray}{25.93}\,/\,\textcolor{gray}{7.90} & \textcolor{gray}{26.97}\,/\,\textcolor{gray}{8.32} & 26.08\,/\,\textcolor{gray}{12.75} & \textcolor{gray}{51.02}\,/\,\textcolor{gray}{3.08} \\
 & SRNO & 33.92\,/\,2.58 & 31.48\,/\,4.42 & 39.99\,/\,\textcolor{gray}{2.78} & \textcolor{gray}{26.01}\,/\,\textcolor{gray}{8.00} & \textcolor{gray}{27.05}\,/\,\textcolor{gray}{8.45} & 26.23\,/\,12.72 & \textcolor{gray}{50.44}\,/\,\textcolor{gray}{3.13} \\
 & GaussianSR & 33.72\,/\,\textcolor{gray}{2.77} & \textcolor{gray}{30.46}\,/\,\textcolor{gray}{6.78} & \textcolor{gray}{36.17}\,/\,\textcolor{gray}{7.45} & \textcolor{gray}{26.04}\,/\,\textcolor{gray}{7.86} & \textcolor{gray}{27.03}\,/\,\textcolor{gray}{8.34} & \textcolor{gray}{25.93}\,/\,\textcolor{gray}{12.94} & \textcolor{gray}{47.26}\,/\,\textcolor{gray}{5.84} \\
 & \cellcolor{oursrow}\ours{} & \cellcolor{oursrow}\textbf{35.24}\,/\,\textbf{2.25} & \cellcolor{oursrow}\textbf{33.01}\,/\,\textbf{3.84} & \cellcolor{oursrow}\textbf{40.97}\,/\,\textbf{2.65} & \cellcolor{oursrow}\textbf{26.70}\,/\,\textbf{7.04} & \cellcolor{oursrow}\textbf{27.73}\,/\,\textbf{7.79} & \cellcolor{oursrow}\textbf{26.92}\,/\,\textbf{12.36} & \cellcolor{oursrow}\textbf{51.25}\,/\,\textbf{3.00} \\
\bottomrule
\end{tabular}
\end{adjustbox}
\end{table}

The figures of this appendix choose their crops and zoom windows by fixed rules that look at no reconstruction. Each dataset contributes the $256{\times}256$ crop with the median bicubic PSNR at $\times4$ among its candidates, which are the center crops of the test images for ARAD, CAVE and Harvard, the crops of Table~\ref{tab:main} for Pavia~U and Chikusei, and non-overlapping crops for Pavia~C and Botswana. The zoom window is the one with the strongest gradients of the band-averaged ground truth.

\begin{figure}[!htbp]
\centering
\includegraphics[width=\linewidth]{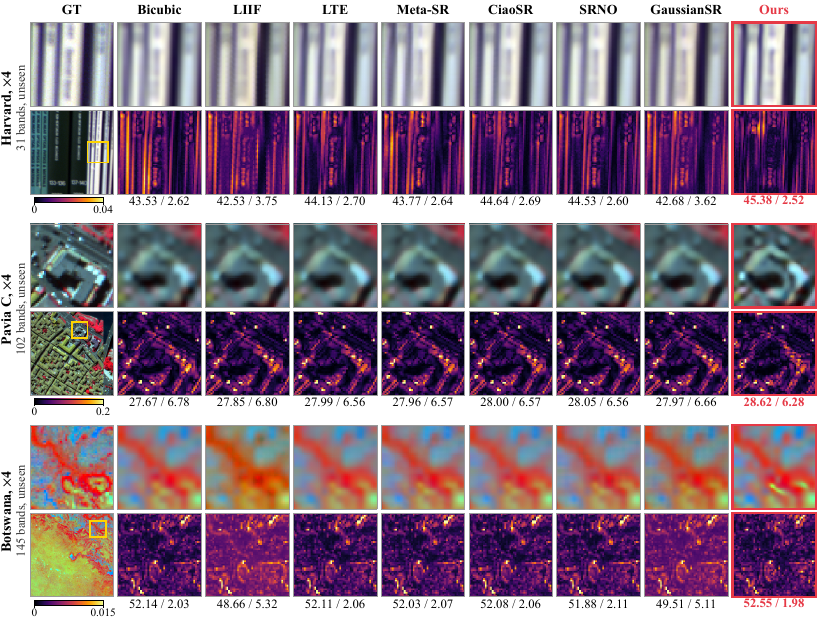}
\caption{Zero-shot reconstructions at $\times$4 on the other three unseen datasets, laid out as Figure~\ref{fig:qual} without the spectra. Pavia~C and Botswana use a near-infrared composite of about 810, 650 and 550~nm. Numbers are PSNR / SAM on the crop.}
\label{fig:appx_unseen}
\end{figure}

Figure~\ref{fig:appx_unseen} extends Figure~\ref{fig:qual} to the other three unseen datasets at $\times4$. \ours{} has the highest PSNR and the lowest SAM on all three crops.

\FloatBarrier
\subsection{SSIM and SAM}
\label{app:ssim}
\label{app:ssim_sam}

Tables~\ref{tab:ssim_sam_chik} and~\ref{tab:ssim_sam_arad} give SSIM and SAM on Chikusei and on the in-domain ARAD under the models, protocol and twelve factors of Table~\ref{tab:main}, the counterparts of Table~\ref{tab:ssim_sam}. On both datasets \ours{} has the highest SSIM and the lowest SAM at all twelve factors. On Chikusei, every baseline has a higher SAM than bicubic interpolation from $\times24$ on and a lower SSIM from $\times36$ on, while \ours{} stays better than interpolation in both.

Table~\ref{tab:ssim} gives SSIM under the target-trained usages of Appendix~\ref{app:routes}, with the better of training from scratch and adapter tuning for each baseline, at $\times4$, $\times8$ and $\times12$, the factors measured for every configuration. The zero-shot \ours{} has the highest SSIM at $\times4$ and $\times8$ on both scenes, and at $\times12$ all methods and bicubic interpolation lie within 0.02 of each other.

\begin{table}[!htbp]
\centering
\caption{SSIM$\uparrow$ and SAM$\downarrow$ (degrees) on Chikusei at the factors of Table~\ref{tab:main}; the Chikusei counterpart of Table~\ref{tab:ssim_sam}.}
\label{tab:ssim_sam_chik}
\footnotesize\renewcommand{\arraystretch}{1.12}\setlength{\tabcolsep}{2.6pt}
\begin{adjustbox}{max width=\linewidth,center}
\begin{tabular}{@{}cc*{12}{c}@{}}
\toprule
\multirow{2}{*}{Metric} & \multirow{2}{*}{Method} & \multicolumn{3}{c}{In-range integers} & \multicolumn{4}{c}{In-range fractional} & \multicolumn{5}{c}{Out-of-range extrapolation} \\
\cmidrule(lr){3-5}\cmidrule(lr){6-9}\cmidrule(lr){10-14}
 & & $\times$2 & $\times$4 & $\times$8 & $\times$2.5 & $\times$3.7 & $\times$5.3 & $\times$7.6 & $\times$12 & $\times$16 & $\times$24 & $\times$36 & $\times$48 \\
\midrule
\multirow{8}{*}{\rotatebox{90}{SSIM$\uparrow$}} & Bicubic & 0.929 & 0.753 & 0.587 & 0.889 & 0.776 & 0.668 & 0.596 & 0.536 & 0.515 & \underline{0.500} & \underline{0.485} & \underline{0.477} \\
 & LIIF & \textcolor{gray}{0.924} & \textcolor{gray}{0.752} & 0.587 & \textcolor{gray}{0.887} & \textcolor{gray}{0.775} & 0.669 & 0.597 & \textcolor{gray}{0.535} & \textcolor{gray}{0.514} & \textcolor{gray}{0.496} & \textcolor{gray}{0.482} & \textcolor{gray}{0.473} \\
 & LTE & 0.936 & 0.767 & 0.598 & 0.900 & 0.791 & 0.684 & \underline{0.608} & \underline{0.542} & \underline{0.520} & \underline{0.500} & \textcolor{gray}{0.484} & \textcolor{gray}{0.475} \\
 & Meta-SR & \underline{0.938} & 0.766 & 0.595 & 0.900 & 0.789 & 0.681 & 0.605 & 0.540 & 0.518 & \textcolor{gray}{0.498} & \textcolor{gray}{0.482} & \textcolor{gray}{0.473} \\
 & CiaoSR & 0.936 & 0.766 & 0.597 & 0.899 & 0.789 & 0.683 & 0.607 & 0.541 & 0.519 & \textcolor{gray}{0.498} & \textcolor{gray}{0.481} & \textcolor{gray}{0.472} \\
 & SRNO & \underline{0.938} & \underline{0.770} & \underline{0.599} & \underline{0.902} & \underline{0.793} & \underline{0.686} & \underline{0.608} & \underline{0.542} & \underline{0.520} & \underline{0.500} & \textcolor{gray}{0.484} & \textcolor{gray}{0.473} \\
 & GaussianSR & 0.932 & 0.762 & 0.592 & 0.895 & 0.785 & 0.678 & 0.601 & 0.537 & \textcolor{gray}{0.515} & \textcolor{gray}{0.495} & \textcolor{gray}{0.479} & \textcolor{gray}{0.470} \\
 & \cellcolor{oursrow}\ours{} & \cellcolor{oursrow}\textbf{0.957} & \cellcolor{oursrow}\textbf{0.813} & \cellcolor{oursrow}\textbf{0.631} & \cellcolor{oursrow}\textbf{0.928} & \cellcolor{oursrow}\textbf{0.834} & \cellcolor{oursrow}\textbf{0.731} & \cellcolor{oursrow}\textbf{0.643} & \cellcolor{oursrow}\textbf{0.558} & \cellcolor{oursrow}\textbf{0.529} & \cellcolor{oursrow}\textbf{0.503} & \cellcolor{oursrow}\textbf{0.487} & \cellcolor{oursrow}\textbf{0.479} \\
\midrule
\multirow{8}{*}{\rotatebox{90}{SAM$\downarrow$ (deg)}} & Bicubic & 2.10 & 3.95 & 5.79 & 2.58 & 3.73 & 4.76 & 5.65 & 6.91 & 7.64 & \underline{8.46} & \underline{9.42} & \underline{10.15} \\
 & LIIF & \textcolor{gray}{2.31} & \textcolor{gray}{4.02} & \textcolor{gray}{5.90} & \textcolor{gray}{2.74} & \textcolor{gray}{3.80} & \textcolor{gray}{4.81} & \textcolor{gray}{5.74} & \textcolor{gray}{7.12} & \textcolor{gray}{7.90} & \textcolor{gray}{8.81} & \textcolor{gray}{9.81} & \textcolor{gray}{10.57} \\
 & LTE & 1.95 & \underline{3.71} & \underline{5.64} & 2.39 & \underline{3.49} & \underline{4.53} & \underline{5.48} & \underline{6.83} & \underline{7.61} & \textcolor{gray}{8.53} & \textcolor{gray}{9.57} & \textcolor{gray}{10.43} \\
 & Meta-SR & \underline{1.91} & 3.73 & 5.68 & \underline{2.38} & 3.51 & 4.57 & 5.53 & 6.90 & \textcolor{gray}{7.68} & \textcolor{gray}{8.61} & \textcolor{gray}{9.65} & \textcolor{gray}{10.47} \\
 & CiaoSR & 1.95 & 3.73 & 5.65 & 2.41 & 3.51 & 4.55 & 5.50 & 6.86 & 7.63 & \textcolor{gray}{8.64} & \textcolor{gray}{9.80} & \textcolor{gray}{10.60} \\
 & SRNO & 1.93 & \underline{3.71} & 5.65 & \underline{2.38} & \underline{3.49} & 4.54 & 5.49 & 6.85 & 7.64 & \textcolor{gray}{8.56} & \textcolor{gray}{9.64} & \textcolor{gray}{10.45} \\
 & GaussianSR & \textcolor{gray}{2.16} & 3.85 & \textcolor{gray}{5.81} & \textcolor{gray}{2.59} & 3.64 & 4.69 & 5.65 & \textcolor{gray}{7.02} & \textcolor{gray}{7.81} & \textcolor{gray}{8.75} & \textcolor{gray}{9.83} & \textcolor{gray}{10.65} \\
 & \cellcolor{oursrow}\ours{} & \cellcolor{oursrow}\textbf{1.70} & \cellcolor{oursrow}\textbf{3.38} & \cellcolor{oursrow}\textbf{5.29} & \cellcolor{oursrow}\textbf{2.14} & \cellcolor{oursrow}\textbf{3.17} & \cellcolor{oursrow}\textbf{4.18} & \cellcolor{oursrow}\textbf{5.10} & \cellcolor{oursrow}\textbf{6.49} & \cellcolor{oursrow}\textbf{7.26} & \cellcolor{oursrow}\textbf{8.27} & \cellcolor{oursrow}\textbf{9.41} & \cellcolor{oursrow}\textbf{10.09} \\
\bottomrule
\end{tabular}
\end{adjustbox}
\end{table}

\begin{table}[!htbp]
\centering
\caption{SSIM$\uparrow$ and SAM$\downarrow$ (degrees) on the in-domain ARAD at the factors of Table~\ref{tab:main}; the ARAD counterpart of Table~\ref{tab:ssim_sam}.}
\label{tab:ssim_sam_arad}
\footnotesize\renewcommand{\arraystretch}{1.12}\setlength{\tabcolsep}{2.6pt}
\begin{adjustbox}{max width=\linewidth,center}
\begin{tabular}{@{}cc*{12}{c}@{}}
\toprule
\multirow{2}{*}{Metric} & \multirow{2}{*}{Method} & \multicolumn{3}{c}{In-range integers} & \multicolumn{4}{c}{In-range fractional} & \multicolumn{5}{c}{Out-of-range extrapolation} \\
\cmidrule(lr){3-5}\cmidrule(lr){6-9}\cmidrule(lr){10-14}
 & & $\times$2 & $\times$4 & $\times$8 & $\times$2.5 & $\times$3.7 & $\times$5.3 & $\times$7.6 & $\times$12 & $\times$16 & $\times$24 & $\times$36 & $\times$48 \\
\midrule
\multirow{8}{*}{\rotatebox{90}{SSIM$\uparrow$}} & Bicubic & 0.978 & 0.915 & 0.841 & 0.963 & 0.924 & 0.883 & 0.847 & 0.807 & 0.790 & 0.768 & 0.748 & 0.731 \\
 & LIIF & 0.981 & 0.923 & 0.849 & 0.967 & 0.931 & 0.891 & 0.855 & 0.812 & 0.792 & 0.769 & \textcolor{gray}{0.747} & \textcolor{gray}{0.730} \\
 & LTE & 0.982 & 0.926 & \underline{0.853} & \underline{0.970} & 0.934 & 0.895 & \underline{0.860} & \underline{0.817} & \underline{0.796} & \underline{0.773} & 0.749 & \underline{0.732} \\
 & Meta-SR & \underline{0.983} & 0.926 & \underline{0.853} & \underline{0.970} & 0.934 & 0.895 & 0.859 & 0.816 & \underline{0.796} & 0.772 & 0.748 & \textcolor{gray}{0.730} \\
 & CiaoSR & \underline{0.983} & 0.926 & \underline{0.853} & \underline{0.970} & 0.934 & 0.895 & 0.859 & 0.815 & 0.795 & 0.770 & \textcolor{gray}{0.746} & \textcolor{gray}{0.728} \\
 & SRNO & \underline{0.983} & \underline{0.928} & \underline{0.853} & \underline{0.970} & \underline{0.936} & \underline{0.896} & 0.859 & 0.815 & 0.795 & 0.772 & \underline{0.750} & 0.731 \\
 & GaussianSR & 0.982 & 0.926 & 0.852 & 0.969 & 0.934 & 0.894 & 0.858 & 0.814 & 0.793 & 0.769 & \textcolor{gray}{0.743} & \textcolor{gray}{0.723} \\
 & \cellcolor{oursrow}\ours{} & \cellcolor{oursrow}\textbf{0.989} & \cellcolor{oursrow}\textbf{0.944} & \cellcolor{oursrow}\textbf{0.874} & \cellcolor{oursrow}\textbf{0.980} & \cellcolor{oursrow}\textbf{0.951} & \cellcolor{oursrow}\textbf{0.916} & \cellcolor{oursrow}\textbf{0.880} & \cellcolor{oursrow}\textbf{0.836} & \cellcolor{oursrow}\textbf{0.813} & \cellcolor{oursrow}\textbf{0.786} & \cellcolor{oursrow}\textbf{0.758} & \cellcolor{oursrow}\textbf{0.741} \\
\midrule
\multirow{8}{*}{\rotatebox{90}{SAM$\downarrow$ (deg)}} & Bicubic & 0.91 & 1.74 & 2.72 & 1.15 & 1.64 & 2.13 & 2.62 & 3.38 & 3.84 & 4.52 & 5.43 & 6.22 \\
 & LIIF & \textcolor{gray}{1.22} & \textcolor{gray}{1.94} & \textcolor{gray}{2.89} & \textcolor{gray}{1.42} & \textcolor{gray}{1.85} & \textcolor{gray}{2.31} & \textcolor{gray}{2.80} & \textcolor{gray}{3.57} & \textcolor{gray}{4.06} & \textcolor{gray}{4.79} & \textcolor{gray}{5.76} & \textcolor{gray}{6.51} \\
 & LTE & 0.90 & 1.64 & 2.58 & 1.10 & \underline{1.54} & 2.01 & 2.50 & 3.24 & 3.70 & \underline{4.39} & \underline{5.31} & \underline{6.06} \\
 & Meta-SR & \underline{0.87} & 1.66 & 2.63 & \underline{1.09} & 1.56 & 2.04 & 2.53 & 3.29 & 3.77 & 4.46 & 5.39 & 6.11 \\
 & CiaoSR & 0.88 & 1.64 & \underline{2.57} & 1.10 & \underline{1.54} & \underline{2.00} & \underline{2.48} & \underline{3.23} & \underline{3.69} & \underline{4.39} & 5.33 & 6.12 \\
 & SRNO & \underline{0.87} & \underline{1.63} & 2.58 & \underline{1.09} & \underline{1.54} & \underline{2.00} & 2.49 & 3.24 & 3.71 & 4.40 & 5.33 & 6.14 \\
 & GaussianSR & \textcolor{gray}{1.16} & \textcolor{gray}{1.81} & \textcolor{gray}{2.77} & \textcolor{gray}{1.31} & \textcolor{gray}{1.72} & \textcolor{gray}{2.18} & \textcolor{gray}{2.68} & \textcolor{gray}{3.45} & \textcolor{gray}{3.94} & \textcolor{gray}{4.67} & \textcolor{gray}{5.69} & \textcolor{gray}{6.50} \\
 & \cellcolor{oursrow}\ours{} & \cellcolor{oursrow}\textbf{0.68} & \cellcolor{oursrow}\textbf{1.38} & \cellcolor{oursrow}\textbf{2.25} & \cellcolor{oursrow}\textbf{0.88} & \cellcolor{oursrow}\textbf{1.29} & \cellcolor{oursrow}\textbf{1.71} & \cellcolor{oursrow}\textbf{2.16} & \cellcolor{oursrow}\textbf{2.86} & \cellcolor{oursrow}\textbf{3.32} & \cellcolor{oursrow}\textbf{3.98} & \cellcolor{oursrow}\textbf{4.98} & \cellcolor{oursrow}\textbf{5.79} \\
\bottomrule
\end{tabular}
\end{adjustbox}
\end{table}

\begin{table}[!htbp]
\centering
\caption{SSIM$\uparrow$ under the target-trained configurations of Table~\ref{tab:routes_full}, at the three factors for which every configuration was measured. $^\dagger$: training from scratch was better than adapter tuning. \textbf{Bold}: best; \underline{underline}: second best.}
\label{tab:ssim}
\footnotesize\renewcommand{\arraystretch}{1.12}\setlength{\tabcolsep}{4pt}
\begin{adjustbox}{max width=\linewidth,center}
\begin{tabular}{@{}l ccc ccc@{}}
\toprule
\multirow{2}{*}{Method} & \multicolumn{3}{c}{Pavia~U ($103$ bands)} & \multicolumn{3}{c}{Chikusei ($128$ bands)} \\
\cmidrule(lr){2-4}\cmidrule(lr){5-7}
 & $\times$4 & $\times$8 & $\times$12 & $\times$4 & $\times$8 & $\times$12 \\
\midrule
Bicubic & 0.785 & 0.641 & 0.591 & 0.792 & 0.648 & 0.594 \\
LIIF & 0.794 & 0.646 & 0.590 & 0.810$^\dagger$ & 0.661$^\dagger$ & 0.600$^\dagger$ \\
LTE & 0.809 & 0.655 & 0.593 & 0.818 & \underline{0.672} & \textbf{0.608} \\
Meta-SR & 0.811 & 0.656 & \textbf{0.595} & 0.816 & 0.668 & \underline{0.605} \\
CiaoSR & 0.808 & 0.651 & 0.589 & 0.817 & 0.670 & 0.604 \\
SRNO & \underline{0.812} & \underline{0.657} & 0.593 & \underline{0.819} & 0.671 & 0.603 \\
GaussianSR & 0.806 & 0.652 & 0.591 & 0.813$^\dagger$ & 0.656 & 0.593$^\dagger$ \\
\cellcolor{oursrow}\ours{} & \cellcolor{oursrow}\textbf{0.824} & \cellcolor{oursrow}\textbf{0.669} & \cellcolor{oursrow}\underline{0.594} & \cellcolor{oursrow}\textbf{0.843} & \cellcolor{oursrow}\textbf{0.683} & \cellcolor{oursrow}0.602 \\
\bottomrule
\end{tabular}
\end{adjustbox}
\end{table}

\begin{figure}[!htbp]
\centering
\includegraphics[width=\linewidth]{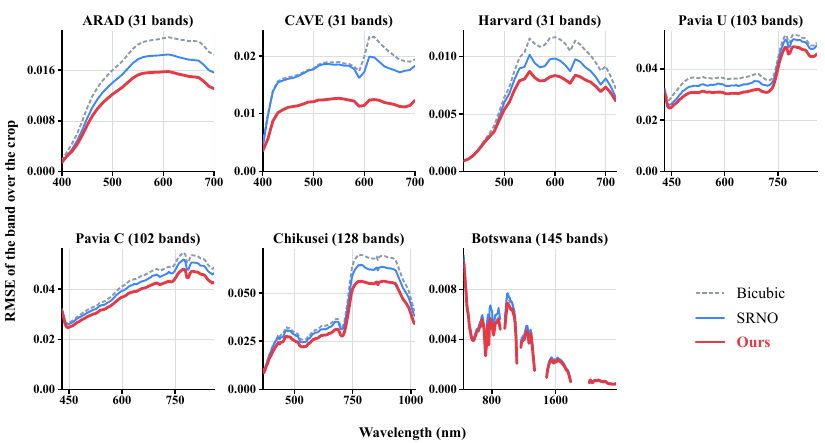}
\caption{RMSE of every band over the crop at $\times$4 on the seven datasets, with the crops chosen as in Appendix~\ref{app:breadth}.}
\label{fig:appx_spectra}
\end{figure}

Figure~\ref{fig:appx_spectra} plots the RMSE of every band over the crops of the seven datasets. \ours{} has the lowest error of all methods in every band on five datasets and in 94\% and 84\% of the bands on ARAD and Harvard, including the near-infrared bands of Pavia~U, Pavia~C and Chikusei and the short-wave infrared bands of Botswana, which lie outside the 400 to 700~nm of the training sensor. Taking the 10\% of pixels with the strongest gradients of the band-averaged ground truth as edge pixels, among the tenth of edge pixels where bicubic interpolation errs most, the spectrum of \ours{} is closer to the ground truth than that of SRNO at 76 to 90\% of the pixels.

\FloatBarrier
\subsection{Reconstructions at \texorpdfstring{$\times$24}{x24}}
\label{app:x24}

\begin{figure}[!htbp]
\centering
\includegraphics[width=\linewidth]{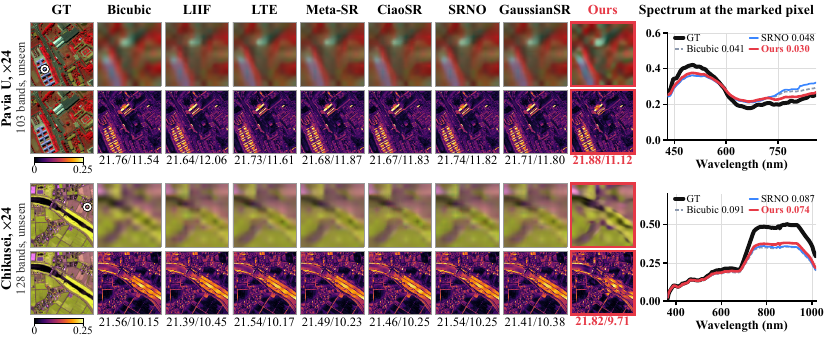}
\caption{Zero-shot reconstructions at $\times$24 on Pavia~U and Chikusei, laid out as Figure~\ref{fig:qual}. The input has $11{\times}11$ pixels, so the zoom shows the whole crop. Numbers are PSNR / SAM on the crop, and the spectrum is taken at the edge pixel with the median bicubic RMSE.}
\label{fig:appx_x24}
\end{figure}

Figure~\ref{fig:appx_x24} compares the reconstructions at $\times24$, outside the training range, under the protocol of Table~\ref{tab:main}, where the input has only $11{\times}11$ pixels. The baselines reproduce the blur of bicubic interpolation, while \ours{} keeps the building rows of Pavia~U and the river bank of Chikusei sharper, with the highest PSNR and the lowest SAM on both crops.

\section{Analysis of \ours{}}
\label{app:analysis}

\subsection{Testing the Band-Shared Operator Prior}
\label{app:prior}

Table~\ref{tab:prior} tests the band-shared operator prior of Section~\ref{sec:method:prior} without any network. The images, crops and degradation are those of the seven-dataset evaluation (Section~\ref{sec:exp:setup} and Table~\ref{tab:ablation_full}). For every image and factor $s \in \{2, 4\}$, we describe $Y_b$ in every window of $16 \times 16$ low-resolution pixels by Eq.~(\ref{eq:prior}), with a zero-sum $5 \times 5$ operator $\mathbf{k}$ of $24$ free coefficients. The low-resolution pixels of a window are split into two halves in a checkerboard pattern; $\mathbf{k}$ is fitted by least squares with a small ridge term on the high-resolution pixels of one half and evaluated on the other half, so no high-resolution pixel is used for both.

A shared operator is fitted on a subset of the bands, either one band, four interior bands evenly spaced over the spectrum, or the first half of the spectrum, and is applied unchanged to all bands outside the subset. As the reference, a per-band operator is fitted for every band on the same pixels. The gain of an operator is $10\log_{10}$ of the ratio between the squared errors of bicubic interpolation and of the operator, where the errors are accumulated over all windows and evaluated bands of an image and the gains are averaged over the images.

An operator fitted on four bands is at least as accurate as the per-band operators on all seven datasets at both factors, and an operator fitted on the first half of the spectrum serves the second half equally well on all datasets except CAVE. These two operators are fitted on more pixels than a per-band operator, which explains their slight advantage. The one-band operator is fitted on as many pixels as a per-band operator, so its column separates the sharing from the number of fitting pixels. It loses at most 0.01~dB on six datasets, while on CAVE it loses 0.29~dB at $\times2$ and 0.13~dB at $\times4$, so the operator should be determined from several bands spread over the spectrum, which the fixed band positions of CSM provide. The per-band gains, 0.2 to 1.2~dB, come from a linear operator fitted per window and serve only as the reference for sharing. The operators of \ours{} form a finer class, one for every primitive placed with a Gaussian support, and this test covers the coarser class of one operator per window.

\begin{table}[!htbp]
\centering
\caption{Test of the band-shared operator prior at $\times$2 and $\times$4, with no network. Per band: PSNR gain (dB) over bicubic interpolation of the operators fitted on each band. Other columns: gain of one shared operator, fitted on one band, four bands or the first half of the spectrum and applied to the remaining bands, minus that of the per-band operators on the same bands. Gray: the shared operator is less accurate.}
\label{tab:prior}
\footnotesize\renewcommand{\arraystretch}{1.12}\setlength{\tabcolsep}{4pt}
\begin{adjustbox}{max width=\linewidth,center}
\begin{tabular}{@{}l cccc cccc@{}}
\toprule
 & \multicolumn{4}{c}{$\times$2} & \multicolumn{4}{c}{$\times$4} \\
\cmidrule(lr){2-5}\cmidrule(lr){6-9}
Dataset & Per band & 1 band & 4 bands & First half & Per band & 1 band & 4 bands & First half \\
\midrule
ARAD (31) & 1.19 & \textcolor{gray}{$-$0.01} & $+$0.03 & $+$0.01 & 0.49 & $+$0.01 & $+$0.02 & $+$0.01 \\
CAVE (31) & 0.73 & \textcolor{gray}{$-$0.29} & $+$0.07 & \textcolor{gray}{$-$0.10} & 0.33 & \textcolor{gray}{$-$0.13} & $+$0.04 & \textcolor{gray}{$-$0.06} \\
Harvard (31) & 0.59 & $+$0.06 & $+$0.10 & $+$0.10 & 0.57 & $+$0.01 & $+$0.03 & $+$0.03 \\
Pavia~U (103) & 0.64 & $+$0.04 & $+$0.17 & $+$0.05 & 0.31 & $+$0.04 & $+$0.08 & $+$0.04 \\
Pavia~C (102) & 0.67 & $+$0.04 & $+$0.14 & $+$0.08 & 0.31 & $+$0.01 & $+$0.06 & $+$0.03 \\
Chikusei (128) & 0.82 & $+$0.00 & $+$0.09 & $+$0.06 & 0.34 & $+$0.02 & $+$0.06 & $+$0.03 \\
Botswana (145) & 0.37 & $+$0.01 & $+$0.12 & $+$0.12 & 0.17 & $+$0.01 & $+$0.04 & $+$0.04 \\
\bottomrule
\end{tabular}
\end{adjustbox}
\end{table}

\subsection{Output Path and Operator Constraint}
\label{app:ablation}

Table~\ref{tab:ablation_full} compares outputs of the network at $\times2$, $\times4$ and $\times8$ on all seven datasets, with the backbone and training shared. The spectral-value column outputs the spectral values outright and combines no observed pixels, and it falls below bicubic interpolation in PSNR in 19 of the 21 rows and below \ours{} in all 21. The free-kernel column drops the zero-sum constraint of the operator, the counterpart of the w/o zero-sum row of Table~\ref{tab:ablation}. It stays above bicubic interpolation in every row and below \ours{} in PSNR in 20 of the 21 rows.

\begin{table}[!htbp]
\centering
\caption{Output path and operator constraint at $\times$2, $\times$4 and $\times$8 on seven datasets (PSNR$\uparrow$ / SAM$\downarrow$). Every model, \ours{} included, is trained on the 900 ARAD images with the same schedule, shorter than that of the main model, so the \ours{} column differs from the main model in Table~\ref{tab:breadth_full}. Remote-sensing scenes are scored on $64{\times}64$ test tiles, not on the $256{\times}256$ crops of Table~\ref{tab:main}. Gray: worse than bicubic interpolation. \textbf{Bold}: best in the row.}
\label{tab:ablation_full}
\footnotesize\renewcommand{\arraystretch}{1.12}\setlength{\tabcolsep}{3pt}
\begin{adjustbox}{max width=\linewidth,center}
\begin{tabular}{@{}l l *{4}{c}@{}}
\toprule
 & \multirow{2}{*}{Dataset} & \multirow{2}{*}{Bicubic} & Predicted & Free & \multirow{2}{*}{\ours{}} \\
 & & & spectral values & kernel & \\
\midrule
\multirow{7}{*}{$\times$2} & ARAD (31) & 44.04\,/\,0.91 & \textcolor{gray}{43.98}\,/\,\textcolor{gray}{1.94} & 47.85\,/\,0.71 & \cellcolor{oursrow}\textbf{47.94}\,/\,\textbf{0.69} \\
 & CAVE (31) & 40.83\,/\,1.95 & \textcolor{gray}{35.07}\,/\,\textcolor{gray}{11.63} & 43.81\,/\,1.64 & \cellcolor{oursrow}\textbf{44.00}\,/\,\textbf{1.62} \\
 & Harvard (31) & 47.47\,/\,1.90 & \textcolor{gray}{40.04}\,/\,\textcolor{gray}{8.61} & 48.84\,/\,1.82 & \cellcolor{oursrow}\textbf{48.91}\,/\,\textbf{1.81} \\
 & Pavia~U (103) & 33.49\,/\,3.56 & \textcolor{gray}{32.50}\,/\,\textcolor{gray}{5.79} & 35.26\,/\,3.26 & \cellcolor{oursrow}\textbf{35.28}\,/\,\textbf{3.21} \\
 & Pavia~C (102) & 34.60\,/\,4.60 & \textcolor{gray}{34.03}\,/\,\textcolor{gray}{6.99} & 36.50\,/\,4.25 & \cellcolor{oursrow}\textbf{36.55}\,/\,\textbf{4.21} \\
 & Chikusei (128) & 34.62\,/\,9.46 & \textcolor{gray}{29.26}\,/\,\textcolor{gray}{14.60} & 37.07\,/\,9.19 & \cellcolor{oursrow}\textbf{37.23}\,/\,\textbf{9.16} \\
 & Botswana (145) & 56.31\,/\,1.64 & \textcolor{gray}{40.06}\,/\,\textcolor{gray}{15.20} & 56.82\,/\,1.56 & \cellcolor{oursrow}\textbf{57.04}\,/\,\textbf{1.54} \\
\midrule
\multirow{7}{*}{$\times$4} & ARAD (31) & 37.50\,/\,1.74 & 38.22\,/\,\textcolor{gray}{2.42} & 39.86\,/\,1.45 & \cellcolor{oursrow}\textbf{39.94}\,/\,\textbf{1.42} \\
 & CAVE (31) & 34.88\,/\,3.12 & \textcolor{gray}{32.30}\,/\,\textcolor{gray}{11.96} & 37.22\,/\,2.61 & \cellcolor{oursrow}\textbf{37.37}\,/\,\textbf{2.58} \\
 & Harvard (31) & 43.09\,/\,2.38 & \textcolor{gray}{38.20}\,/\,\textcolor{gray}{8.79} & 44.54\,/\,\textbf{2.29} & \cellcolor{oursrow}\textbf{44.62}\,/\,\textbf{2.29} \\
 & Pavia~U (103) & 29.04\,/\,5.38 & \textcolor{gray}{28.66}\,/\,\textcolor{gray}{7.05} & 29.87\,/\,5.09 & \cellcolor{oursrow}\textbf{29.92}\,/\,\textbf{5.02} \\
 & Pavia~C (102) & 30.05\,/\,6.39 & \textcolor{gray}{29.88}\,/\,\textcolor{gray}{8.16} & 30.95\,/\,6.09 & \cellcolor{oursrow}\textbf{31.04}\,/\,\textbf{6.05} \\
 & Chikusei (128) & 29.10\,/\,11.09 & \textcolor{gray}{26.39}\,/\,\textcolor{gray}{15.42} & 30.53\,/\,10.78 & \cellcolor{oursrow}\textbf{30.69}\,/\,\textbf{10.73} \\
 & Botswana (145) & 53.07\,/\,2.38 & \textcolor{gray}{39.41}\,/\,\textcolor{gray}{15.26} & 53.17\,/\,2.35 & \cellcolor{oursrow}\textbf{53.32}\,/\,\textbf{2.32} \\
\midrule
\multirow{7}{*}{$\times$8} & ARAD (31) & 33.37\,/\,2.72 & 34.04\,/\,\textcolor{gray}{3.13} & 34.88\,/\,2.35 & \cellcolor{oursrow}\textbf{34.96}\,/\,\textbf{2.33} \\
 & CAVE (31) & 30.86\,/\,4.71 & \textcolor{gray}{29.62}\,/\,\textcolor{gray}{12.53} & 32.59\,/\,3.97 & \cellcolor{oursrow}\textbf{32.68}\,/\,\textbf{3.96} \\
 & Harvard (31) & 39.56\,/\,2.76 & \textcolor{gray}{35.83}\,/\,\textcolor{gray}{8.98} & \textbf{40.74}\,/\,\textbf{2.67} & \cellcolor{oursrow}40.68\,/\,\textbf{2.67} \\
 & Pavia~U (103) & 26.12\,/\,7.63 & \textcolor{gray}{25.89}\,/\,\textcolor{gray}{8.90} & 26.66\,/\,7.19 & \cellcolor{oursrow}\textbf{26.69}\,/\,\textbf{7.14} \\
 & Pavia~C (102) & 27.12\,/\,8.18 & \textcolor{gray}{26.91}\,/\,\textcolor{gray}{9.65} & 27.61\,/\,7.92 & \cellcolor{oursrow}\textbf{27.66}\,/\,\textbf{7.89} \\
 & Chikusei (128) & 26.00\,/\,12.73 & \textcolor{gray}{24.27}\,/\,\textcolor{gray}{16.49} & 26.71\,/\,12.48 & \cellcolor{oursrow}\textbf{26.74}\,/\,\textbf{12.47} \\
 & Botswana (145) & 51.22\,/\,3.00 & \textcolor{gray}{38.64}\,/\,\textcolor{gray}{15.44} & 51.23\,/\,2.99 & \cellcolor{oursrow}\textbf{51.29}\,/\,\textbf{2.98} \\
\bottomrule
\end{tabular}
\end{adjustbox}
\end{table}

\begin{figure}[!htbp]
\centering
\includegraphics[width=0.95\linewidth]{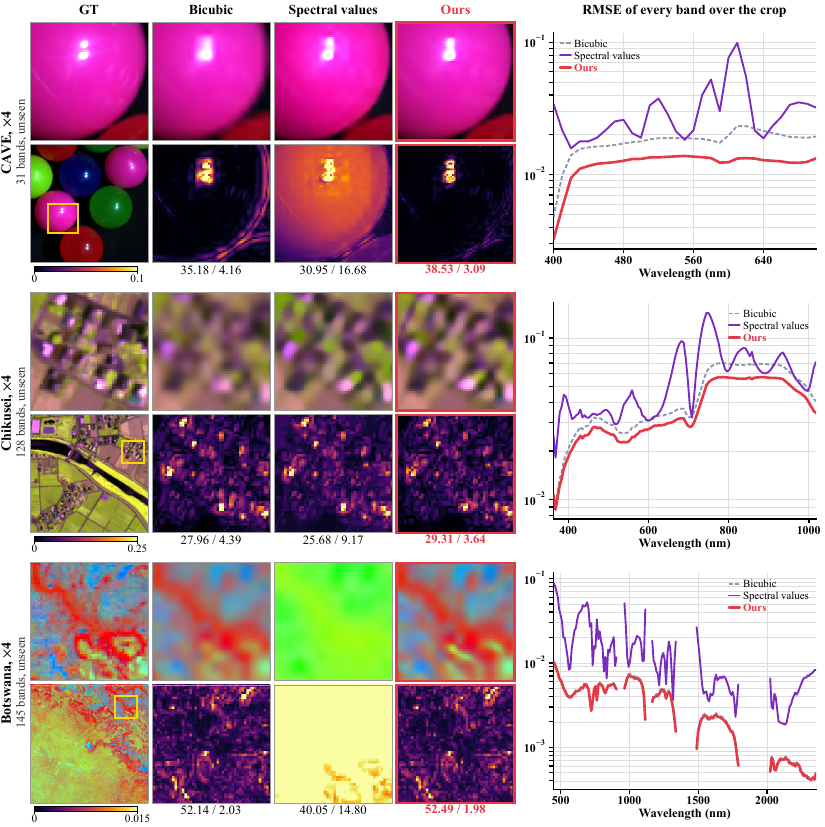}
\caption{Output paths at $\times$4 with the models of Table~\ref{tab:ablation_full}. Spectral values: the network outputs the value of every band. The right column gives the RMSE of every band over the crop on a log scale. Numbers are PSNR / SAM on the crop.}
\label{fig:appx_paths}
\end{figure}

Figure~\ref{fig:appx_paths} compares outputting the spectral values with the operators, both from Table~\ref{tab:ablation_full}, on three unseen datasets at $\times4$. Outputting the spectral values outright shifts the colors of whole regions on CAVE and fails on the $145$-band Botswana, where its error exceeds the color scale everywhere, and its error is above that of the operators in every band of all three crops.

\subsection{Reference Positions and Operator Window}
\label{app:positions}
\label{app:window}

Table~\ref{tab:positions} varies the number $M$ of reference positions from 1 to 62. Each model is retrained on the 900 ARAD images and evaluated on the $256{\times}256$ crops of Table~\ref{tab:main}. From $M{=}8$ on, including $M{=}62$, the PSNR on all three datasets stays within 0.1~dB of $M{=}31$. With a single position, ARAD and Chikusei lose 0.41 and 0.55~dB at $\times$4 and remain above bicubic interpolation. Time and peak memory show no monotonic trend in $M$.

Table~\ref{tab:window} varies the operator window from $3{\times}3$ to $9{\times}9$ with the same budget, and its $5{\times}5$ row is the model in the $M{=}31$ row of Table~\ref{tab:positions}. In PSNR, the $7{\times}7$ window stays within 0.04~dB of $5{\times}5$ on all three datasets. The $3{\times}3$ window is 0.10 to 0.25~dB below $5{\times}5$ and keeps 72 to 82\% of the gain of $5{\times}5$ over bicubic interpolation on the two unseen sensors and about 90\% on ARAD. The $9{\times}9$ window is 0.01 to 0.09~dB below $5{\times}5$. The window changes the parameter count by at most 0.4\%, and no larger window is slower than $5{\times}5$.

\begin{table}[!htbp]
\centering
\caption{Number $M$ of reference positions (PSNR$\uparrow$ / SAM$\downarrow$). Every row is trained on the 900 ARAD images with the same schedule, shorter than that of the main model, so the $M{=}31$ row, the setting of the main model, differs from Table~\ref{tab:main}. Time and peak memory on Chikusei.}
\label{tab:positions}
\footnotesize\renewcommand{\arraystretch}{1.12}\setlength{\tabcolsep}{4pt}
\begin{adjustbox}{max width=\linewidth,center}
\begin{tabular}{@{}c ccc cc@{}}
\toprule
$M$ & ARAD ($31$ bands) & Pavia~U ($103$ bands) & Chikusei ($128$ bands) & Time (ms) & Memory (MiB) \\
\midrule
\multicolumn{6}{@{}l}{\textit{$\times$4}} \\
\midrule
Bicubic & 37.50 / 1.74 & 28.13 / 5.16 & 27.60 / 3.95 & -- & -- \\
1 & 39.53 / 1.52 & 28.86 / 4.91 & 28.23 / 3.81 & 15.4 & 490 \\
2 & 39.76 / 1.46 & 28.82 / 4.93 & 28.61 / 3.52 & 18.7 & 524 \\
4 & 39.83 / 1.45 & 28.87 / 4.83 & 28.72 / 3.45 & 23.2 & 596 \\
8 & 39.89 / 1.42 & 28.92 / 4.78 & 28.84 / 3.44 & 23.1 & 595 \\
16 & 39.86 / 1.42 & 28.89 / 4.79 & 28.80 / 3.46 & 21.8 & 559 \\
\cellcolor{oursrow}31 & \cellcolor{oursrow}39.94 / 1.42 & \cellcolor{oursrow}28.91 / 4.81 & \cellcolor{oursrow}28.78 / 3.48 & \cellcolor{oursrow}21.8 & \cellcolor{oursrow}559 \\
62 & 39.91 / 1.42 & 28.92 / 4.79 & 28.80 / 3.46 & 20.9 & 556 \\
\midrule
\multicolumn{6}{@{}l}{\textit{$\times$8}} \\
\midrule
Bicubic & 33.37 / 2.72 & 25.04 / 7.28 & 24.50 / 5.79 & -- & -- \\
1 & 34.72 / 2.45 & 25.65 / 6.76 & 24.88 / 5.65 & 14.0 & 414 \\
2 & 34.84 / 2.38 & 25.54 / 7.00 & 25.03 / 5.44 & 14.1 & 425 \\
4 & 34.84 / 2.35 & 25.61 / 6.81 & 25.08 / 5.37 & 14.2 & 425 \\
8 & 34.91 / 2.33 & 25.58 / 6.76 & 25.13 / 5.39 & 19.2 & 497 \\
16 & 34.91 / 2.32 & 25.59 / 6.78 & 25.11 / 5.43 & 15.2 & 467 \\
\cellcolor{oursrow}31 & \cellcolor{oursrow}34.96 / 2.33 & \cellcolor{oursrow}25.58 / 6.81 & \cellcolor{oursrow}25.09 / 5.42 & \cellcolor{oursrow}15.1 & \cellcolor{oursrow}468 \\
62 & 34.94 / 2.32 & 25.59 / 6.80 & 25.10 / 5.41 & 14.0 & 426 \\
\bottomrule
\end{tabular}
\end{adjustbox}
\end{table}

\begin{table}[!htbp]
\centering
\caption{Operator window (PSNR$\uparrow$ / SAM$\downarrow$). Every row is trained on the 900 ARAD images with the same schedule, shorter than that of the main model, so the $5{\times}5$ row, the setting of the main model, differs from Table~\ref{tab:main}. Time and peak memory on Chikusei.}
\label{tab:window}
\footnotesize\renewcommand{\arraystretch}{1.12}\setlength{\tabcolsep}{4pt}
\begin{adjustbox}{max width=\linewidth,center}
\begin{tabular}{@{}c ccc cc@{}}
\toprule
Window & ARAD ($31$ bands) & Pavia~U ($103$ bands) & Chikusei ($128$ bands) & Time (ms) & Memory (MiB) \\
\midrule
\multicolumn{6}{@{}l}{\textit{$\times$4}} \\
\midrule
Bicubic & 37.50 / 1.74 & 28.13 / 5.16 & 27.60 / 3.95 & -- & -- \\
$3{\times}3$ & 39.70 / 1.49 & 28.69 / 5.00 & 28.53 / 3.56 & 32.1 & 684 \\
\cellcolor{oursrow}$5{\times}5$ & \cellcolor{oursrow}39.94 / 1.42 & \cellcolor{oursrow}28.91 / 4.81 & \cellcolor{oursrow}28.78 / 3.48 & \cellcolor{oursrow}21.8 & \cellcolor{oursrow}559 \\
$7{\times}7$ & 39.95 / 1.42 & 28.92 / 4.78 & 28.82 / 3.44 & 21.2 & 605 \\
$9{\times}9$ & 39.85 / 1.44 & 28.82 / 4.90 & 28.73 / 3.52 & 21.8 & 666 \\
\midrule
\multicolumn{6}{@{}l}{\textit{$\times$8}} \\
\midrule
Bicubic & 33.37 / 2.72 & 25.04 / 7.28 & 24.50 / 5.79 & -- & -- \\
$3{\times}3$ & 34.81 / 2.37 & 25.48 / 7.05 & 24.98 / 5.48 & 18.4 & 489 \\
\cellcolor{oursrow}$5{\times}5$ & \cellcolor{oursrow}34.96 / 2.33 & \cellcolor{oursrow}25.58 / 6.81 & \cellcolor{oursrow}25.09 / 5.42 & \cellcolor{oursrow}15.1 & \cellcolor{oursrow}468 \\
$7{\times}7$ & 34.94 / 2.31 & 25.58 / 6.80 & 25.10 / 5.39 & 14.9 & 451 \\
$9{\times}9$ & 34.87 / 2.33 & 25.57 / 6.93 & 25.06 / 5.48 & 14.2 & 441 \\
\bottomrule
\end{tabular}
\end{adjustbox}
\end{table}

\FloatBarrier
\subsection{Training on a Different Source Sensor}
\label{app:reverse}

Table~\ref{tab:reverse} swaps the training sensor for the $128$-band Chikusei, keeps everything else fixed, and places this model next to the ARAD-trained main model on the seven datasets at $\times2$, $\times4$ and $\times8$. On the three $31$-band datasets, PSNR and SAM are above bicubic interpolation at $\times2$, $\times4$ and $\times8$ throughout, and so are the remaining datasets except Botswana at $\times8$, where the model ties with interpolation. Transfer holds in both directions, and $31$ is only the choice made for the main model of this paper. The Chikusei-trained model gains less than the ARAD-trained one; its training data is $234$ tiles of a single scene.

\begin{table}[!htbp]
\centering
\caption{Training sensor swapped: the same framework trained on the $128$-band Chikusei (234 tiles, 2000 epochs) against the main model trained on the $31$-band ARAD (900 images, 2000 epochs), PSNR$\uparrow$ / SAM$\downarrow$ on the seven datasets at $\times$2, $\times$4 and $\times$8, single forward pass, no target-domain data for either. ARAD is in-domain for the ARAD-trained model and Chikusei for the Chikusei-trained one. Gray: worse than bicubic interpolation.}
\label{tab:reverse}
\footnotesize\renewcommand{\arraystretch}{1.12}\setlength{\tabcolsep}{4pt}
\begin{adjustbox}{max width=\linewidth,center}
\begin{tabular}{@{}l l *{3}{c}@{}}
\toprule
 & \multirow{2}{*}{Dataset} & \multirow{2}{*}{Bicubic} & \ours{} trained on & \ours{} trained on \\
 & & & ARAD ($31$ bands) & Chikusei ($128$ bands) \\
\midrule
\multirow{7}{*}{$\times$2} & ARAD (31) & 44.04\,/\,0.91 & 48.41\,/\,0.68 & 46.59\,/\,0.79 \\
 & CAVE (31) & 40.83\,/\,1.95 & 44.27\,/\,1.59 & 43.12\,/\,1.76 \\
 & Harvard (31) & 47.47\,/\,1.90 & 49.02\,/\,1.81 & 48.44\,/\,1.83 \\
 & Pavia~U (103) & 33.49\,/\,3.56 & 35.39\,/\,3.19 & 35.11\,/\,3.25 \\
 & Pavia~C (102) & 34.60\,/\,4.60 & 36.67\,/\,4.20 & 36.33\,/\,4.30 \\
 & Chikusei (128) & 34.62\,/\,9.46 & 37.49\,/\,9.13 & 37.68\,/\,9.01 \\
 & Botswana (145) & 56.31\,/\,1.64 & 57.03\,/\,1.55 & 57.02\,/\,1.55 \\
\midrule
\multirow{7}{*}{$\times$4} & ARAD (31) & 37.50\,/\,1.74 & 40.33\,/\,1.38 & 38.99\,/\,1.60 \\
 & CAVE (31) & 34.88\,/\,3.12 & 37.66\,/\,2.53 & 36.33\,/\,2.87 \\
 & Harvard (31) & 43.09\,/\,2.38 & 44.85\,/\,2.28 & 43.93\,/\,2.32 \\
 & Pavia~U (103) & 29.04\,/\,5.38 & 29.98\,/\,4.91 & 29.72\,/\,5.17 \\
 & Pavia~C (102) & 30.05\,/\,6.39 & 31.11\,/\,5.97 & 30.82\,/\,6.20 \\
 & Chikusei (128) & 29.10\,/\,11.09 & 30.92\,/\,10.65 & 31.39\,/\,10.37 \\
 & Botswana (145) & 53.07\,/\,2.38 & 53.26\,/\,2.34 & 53.22\,/\,2.35 \\
\midrule
\multirow{7}{*}{$\times$8} & ARAD (31) & 33.37\,/\,2.72 & 35.24\,/\,2.25 & 34.33\,/\,2.59 \\
 & CAVE (31) & 30.86\,/\,4.71 & 33.01\,/\,3.84 & 31.82\,/\,4.38 \\
 & Harvard (31) & 39.56\,/\,2.76 & 40.97\,/\,2.65 & 40.10\,/\,2.73 \\
 & Pavia~U (103) & 26.12\,/\,7.63 & 26.70\,/\,7.04 & 26.39\,/\,7.44 \\
 & Pavia~C (102) & 27.12\,/\,8.18 & 27.73\,/\,7.79 & 27.44\,/\,8.07 \\
 & Chikusei (128) & 26.00\,/\,12.73 & 26.92\,/\,12.36 & 27.53\,/\,11.95 \\
 & Botswana (145) & 51.22\,/\,3.00 & 51.25\,/\,3.00 & \textcolor{gray}{51.23}\,/\,\textcolor{gray}{3.01} \\
\bottomrule
\end{tabular}
\end{adjustbox}
\end{table}

\FloatBarrier
\subsection{Redefining the Bands}
\label{app:banddef}

Eq.~(\ref{eq:equivariance}) holds for a given $W_s$, and the band definition of a sensor reaches the output of \ours{} only through the prediction of $W_s$ from $Z$ (Section~\ref{sec:method:cofr}). Table~\ref{tab:banddef} redefines the bands of every test image of the seven-dataset evaluation in four ways and runs \ours{} on the redefined input $TX$. It compares the result $f(TX)$ with $Tf(X)$, the reconstruction of the original bands redefined afterwards, which Eq.~(\ref{eq:equivariance}) makes equal to $f(TX)$ if the predicted $W_s$ does not change. The relative change $r = \lVert f(TX) - Tf(X) \rVert / \lVert Tf(X) - U_s TX \rVert$ measures how much of the contribution of the operator field moves, and $\Delta$ compares the two against the redefined ground truth $TY$, without clipping. At $\times4$, the median $r$ is no more than 0.09 for subsets and merges and 0.17 for gains and offsets, and at $\times2$, $\times4$ and $\times8$ the PSNR changes by at most 0.05~dB for every transform and dataset. Redefining the bands changes the predicted operators little and leaves the accuracy of the reconstruction unchanged.

\begin{table}[!htbp]
\centering
\caption{Redefining the input bands at $\times$4. Subset: every other band; merge: means of neighboring pairs; gain and offset: a random per-band gain in $[0.8, 1.25]$ or offset in $[0, 0.05]$. $r$: relative change of the contribution of the operator field, median over images; $\Delta$: PSNR of $f(TX)$ minus that of $Tf(X)$, both against $TY$ (dB).}
\label{tab:banddef}
\footnotesize\renewcommand{\arraystretch}{1.12}\setlength{\tabcolsep}{4pt}
\begin{adjustbox}{max width=\linewidth,center}
\begin{tabular}{@{}l cc cc cc cc@{}}
\toprule
\multirow{2}{*}{Dataset} & \multicolumn{2}{c}{Subset} & \multicolumn{2}{c}{Merge} & \multicolumn{2}{c}{Gain} & \multicolumn{2}{c}{Offset} \\
\cmidrule(lr){2-3}\cmidrule(lr){4-5}\cmidrule(lr){6-7}\cmidrule(lr){8-9}
 & $r$ & $\Delta$ & $r$ & $\Delta$ & $r$ & $\Delta$ & $r$ & $\Delta$ \\
\midrule
ARAD (31) & 0.008 & 0.00 & 0.049 & $-$0.01 & 0.060 & 0.00 & 0.089 & $-$0.02 \\
CAVE (31) & 0.021 & $+$0.01 & 0.064 & $-$0.01 & 0.063 & 0.00 & 0.071 & $+$0.01 \\
Harvard (31) & 0.060 & 0.00 & 0.090 & $+$0.01 & 0.064 & 0.00 & 0.104 & $-$0.03 \\
Pavia~U (103) & 0.007 & 0.00 & 0.038 & $+$0.01 & 0.077 & $+$0.02 & 0.160 & $+$0.04 \\
Pavia~C (102) & 0.017 & 0.00 & 0.038 & $+$0.01 & 0.144 & 0.00 & 0.136 & $-$0.03 \\
Chikusei (128) & 0.013 & 0.00 & 0.025 & 0.00 & 0.076 & $-$0.03 & 0.053 & 0.00 \\
Botswana (145) & 0.018 & 0.00 & 0.037 & 0.00 & 0.108 & $-$0.01 & 0.160 & 0.00 \\
\bottomrule
\end{tabular}
\end{adjustbox}
\end{table}

\FloatBarrier
\subsection{Blur That Grows Along the Spectrum}
\label{app:bandblur}

Every other experiment degrades all bands identically, the setting that the band-shared prior assumes. Table~\ref{tab:bandblur} lets the blur grow along the spectrum. Before the antialiased bicubic downsampling, band $b$ is blurred by a Gaussian of standard deviation $\kappa s (b-1)/(B-1)$ high-resolution pixels, so at $\kappa = 0.5$ the last band carries an extra blur of half a low-resolution pixel, and the models are the ARAD-trained ones of Table~\ref{tab:seven}. Every method loses accuracy as $\kappa$ grows. At $\kappa = 0.5$, \ours{} is above SRNO on every image and tile of the seven datasets, by 0.49 to 1.78~dB on average, and above bicubic interpolation on every dataset, with the lowest SAM of the three methods; SRNO stays below bicubic interpolation on Botswana.

\begin{table}[!htbp]
\centering
\caption{Blur that grows along the spectrum, PSNR$\uparrow$ (dB) at $\times$4. Band $b$ is blurred by a Gaussian of standard deviation $\kappa s (b-1)/(B-1)$ high-resolution pixels before the antialiased bicubic downsampling; $\kappa = 0$ is the setting of Table~\ref{tab:seven}. Gray: below bicubic interpolation. \textbf{Bold}: best for each $\kappa$.}
\label{tab:bandblur}
\footnotesize\renewcommand{\arraystretch}{1.12}\setlength{\tabcolsep}{3pt}
\begin{adjustbox}{max width=\linewidth,center}
\begin{tabular}{@{}l ccc ccc ccc@{}}
\toprule
\multirow{2}{*}{Dataset} & \multicolumn{3}{c}{$\kappa = 0$} & \multicolumn{3}{c}{$\kappa = 0.25$} & \multicolumn{3}{c}{$\kappa = 0.5$} \\
\cmidrule(lr){2-4}\cmidrule(lr){5-7}\cmidrule(lr){8-10}
 & Bicubic & SRNO & \ours{} & Bicubic & SRNO & \ours{} & Bicubic & SRNO & \ours{} \\
\midrule
ARAD (31) & 37.50 & 38.49 & \cellcolor{oursrow}\textbf{40.33} & 37.34 & 38.27 & \cellcolor{oursrow}\textbf{40.16} & 36.85 & 37.57 & \cellcolor{oursrow}\textbf{39.18} \\
CAVE (31) & 34.88 & 35.59 & \cellcolor{oursrow}\textbf{37.66} & 34.72 & 35.38 & \cellcolor{oursrow}\textbf{37.51} & 34.21 & 34.73 & \cellcolor{oursrow}\textbf{36.51} \\
Harvard (31) & 43.09 & 43.53 & \cellcolor{oursrow}\textbf{44.85} & 42.94 & 43.37 & \cellcolor{oursrow}\textbf{44.72} & 42.49 & 42.85 & \cellcolor{oursrow}\textbf{43.99} \\
Pavia~U (103) & 29.04 & 29.40 & \cellcolor{oursrow}\textbf{29.98} & 28.92 & 29.28 & \cellcolor{oursrow}\textbf{29.94} & 28.57 & 28.90 & \cellcolor{oursrow}\textbf{29.56} \\
Pavia~C (102) & 30.05 & 30.43 & \cellcolor{oursrow}\textbf{31.11} & 29.93 & 30.30 & \cellcolor{oursrow}\textbf{31.08} & 29.56 & 29.88 & \cellcolor{oursrow}\textbf{30.63} \\
Chikusei (128) & 29.10 & 29.73 & \cellcolor{oursrow}\textbf{30.92} & 28.98 & 29.58 & \cellcolor{oursrow}\textbf{30.88} & 28.62 & 29.13 & \cellcolor{oursrow}\textbf{30.28} \\
Botswana (145) & 53.07 & \textcolor{gray}{52.81} & \cellcolor{oursrow}\textbf{53.26} & 52.98 & \textcolor{gray}{52.74} & \cellcolor{oursrow}\textbf{53.22} & 52.75 & \textcolor{gray}{52.54} & \cellcolor{oursrow}\textbf{53.03} \\
\bottomrule
\end{tabular}
\end{adjustbox}
\end{table}

\FloatBarrier
\subsection{Test-Time Degradations}
\label{app:degrade}
Table~\ref{tab:degrade} changes only how the test inputs are generated, on the crops of Table~\ref{tab:main}. Noise adds i.i.d.\ Gaussian noise with standard deviation $\sigma$ to the bicubic input. Blur applies an isotropic Gaussian with a standard deviation of $0.25s$, $0.40s$ or $0.55s$ high-resolution pixels and keeps every $s$-th pixel, and area averages $s{\times}s$ blocks.

\begin{table}[!htbp]
\centering
\caption{PSNR$\uparrow$ (dB) under test-time degradations on the crops of Table~\ref{tab:main}. For each dataset: bicubic interpolation, the best of the six baselines, and \ours{}. Gray: below bicubic interpolation; \textbf{bold}: best.}
\label{tab:degrade}
\footnotesize\renewcommand{\arraystretch}{1.12}\setlength{\tabcolsep}{3pt}
\begin{adjustbox}{max width=\linewidth,center}
\begin{tabular}{@{}l *{9}{c}@{}}
\toprule
\multirow{2}{*}{Test input} & \multicolumn{3}{c}{ARAD (31)} & \multicolumn{3}{c}{Pavia~U (103)} & \multicolumn{3}{c}{Chikusei (128)} \\
\cmidrule(lr){2-4}\cmidrule(lr){5-7}\cmidrule(lr){8-10}
 & Bicubic & Best & \ours{} & Bicubic & Best & \ours{} & Bicubic & Best & \ours{} \\
\midrule
\multicolumn{10}{@{}l}{\textit{$\times$4}} \\
\midrule
Bicubic (as in training) & 37.50 & 38.49 & \textbf{40.36} & 28.13 & 28.44 & \textbf{28.92} & 27.60 & 27.96 & \textbf{28.94} \\
Noise $\sigma{=}0.01$ & 35.07 & 36.66 & \textbf{36.75} & 27.92 & 28.23 & \textbf{28.30} & 27.31 & 27.72 & \textbf{27.82} \\
Noise $\sigma{=}0.02$ & 32.65 & 35.21 & \textbf{35.25} & 27.34 & \textbf{28.00} & 27.87 & 26.65 & \textbf{27.42} & 27.26 \\
Noise $\sigma{=}0.03$ & 30.64 & \textbf{33.87} & 30.78 & 26.55 & \textbf{27.68} & 26.64 & 25.85 & \textbf{27.12} & 25.92 \\
Blur $0.25s$ & 36.83 & 37.77 & \textbf{37.79} & 27.59 & \textbf{27.84} & 27.82 & 27.14 & \textbf{27.46} & 27.45 \\
Blur $0.40s$ & 36.34 & 36.97 & \textbf{38.06} & 27.31 & 27.49 & \textbf{27.97} & 26.87 & 27.12 & \textbf{27.83} \\
Blur $0.55s$ & 35.44 & 35.79 & \textbf{36.58} & 26.63 & 26.71 & \textbf{27.10} & 26.16 & 26.30 & \textbf{26.90} \\
Area & 37.53 & 38.76 & \textbf{39.35} & 28.14 & 28.48 & \textbf{28.49} & 27.59 & 27.98 & \textbf{28.36} \\
\midrule
\multicolumn{10}{@{}l}{\textit{$\times$8}} \\
\midrule
Bicubic (as in training) & 33.37 & 33.92 & \textbf{35.26} & 25.04 & 25.19 & \textbf{25.59} & 24.50 & 24.62 & \textbf{25.16} \\
Noise $\sigma{=}0.01$ & 31.98 & 32.78 & \textbf{32.91} & 24.94 & 25.02 & \textbf{25.08} & 24.34 & 24.49 & \textbf{24.57} \\
Noise $\sigma{=}0.02$ & 30.40 & 31.85 & \textbf{31.99} & 24.63 & \textbf{24.80} & 24.76 & 23.97 & \textbf{24.25} & 24.20 \\
Noise $\sigma{=}0.03$ & 28.95 & \textbf{30.95} & 29.07 & 24.17 & \textbf{24.65} & 24.22 & 23.47 & \textbf{24.11} & 23.55 \\
Blur $0.25s$ & 33.27 & \textbf{33.97} & \textbf{33.97} & 24.99 & \textbf{25.15} & 25.12 & 24.36 & \textbf{24.55} & 24.51 \\
Blur $0.40s$ & 32.91 & 33.37 & \textbf{34.52} & 24.68 & 24.82 & \textbf{25.33} & 24.23 & 24.34 & \textbf{24.90} \\
Blur $0.55s$ & 32.21 & 32.47 & \textbf{33.23} & 24.11 & 24.22 & \textbf{24.59} & 23.77 & 23.84 & \textbf{24.23} \\
Area & 33.37 & 34.00 & \textbf{34.48} & 25.05 & 25.20 & \textbf{25.36} & 24.48 & 24.60 & \textbf{24.76} \\
\bottomrule
\end{tabular}
\end{adjustbox}
\end{table}

\subsection{What the Network Predicts}
\label{app:primitives}

\begin{figure}[!htbp]
\centering
\includegraphics[width=\linewidth]{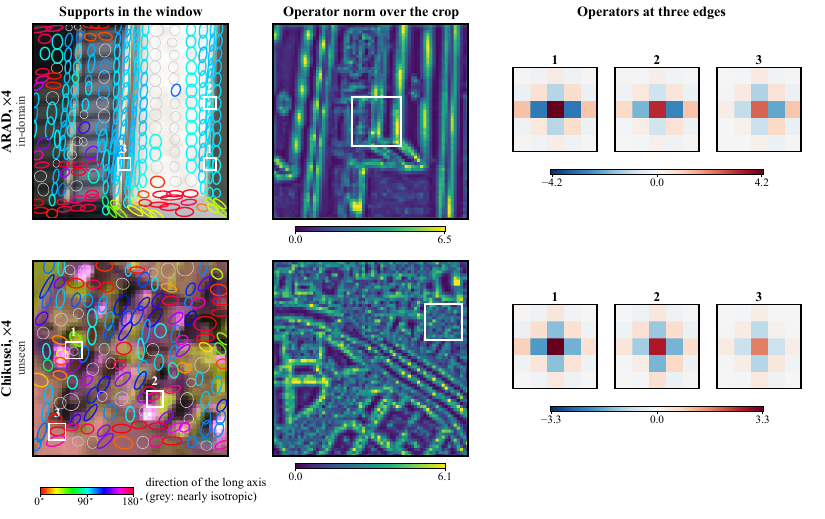}
\caption{What the network predicts at $\times$4. Left: the supports in the window, colored by the direction of the long axis (grey: axis ratio below 1.2). Middle: the norm of every $5{\times}5$ operator over the crop. Right: the three edge operators in the window with the largest norm.}
\label{fig:appx_prims}
\end{figure}

Figure~\ref{fig:appx_prims} shows the primitives and operators that \ours{} predicts at $\times4$ on crops of ARAD and Chikusei chosen as in Appendix~\ref{app:breadth}. At edges the supports stretch along the edge. We call a primitive an edge primitive when the gradient magnitude of the band-averaged ground truth at its low-resolution pixel is in the top 10\%, and a flat primitive when it is in the bottom half. Among the edge primitives with an axis ratio of at least 1.2, the long axis lies within $30^\circ$ of the edge direction for 98.5\% on ARAD and 94.8\% on Chikusei. The median axis ratio is 3.9 at edges and 1.3 in flat regions on ARAD, and 3.8 and 1.6 on Chikusei. The operators concentrate on edges as well, with a median norm of 3.5 at edges and 0.9 in flat regions on ARAD and 2.8 and 1.4 on Chikusei, so the correction to bicubic interpolation is small away from edges.

\FloatBarrier
\section{Inference Cost}
\label{sec:app:efficiency}

\paragraph{Timing in Figure~\ref{fig:teaser}.}
Panels (a,b) profile cached multi-scale inference, whereas (d) times each output independently.
Both use three $32\times32\times145$ Botswana inputs, one NVIDIA A40, batch size~1 and FP32 with TF32 disabled.
For (a,b), each run requests factors $4,6,8,10,12,14,16$, reusing scale-independent features in both models.
After two warm-up bundles, we time five bundles and select the median-total-time run per input, retaining its paired stage durations, then average across the three inputs.
CSM performs feature encoding once but repeats scale conditioning and operator prediction for each output.
The displayed per-output coefficients average over these seven factors and approximate this workload.
Panel (d) averages per-input medians from twelve independent runs after three warm-ups at each factor.

\begin{figure}[!htbp]
\centering
\includegraphics[width=\linewidth]{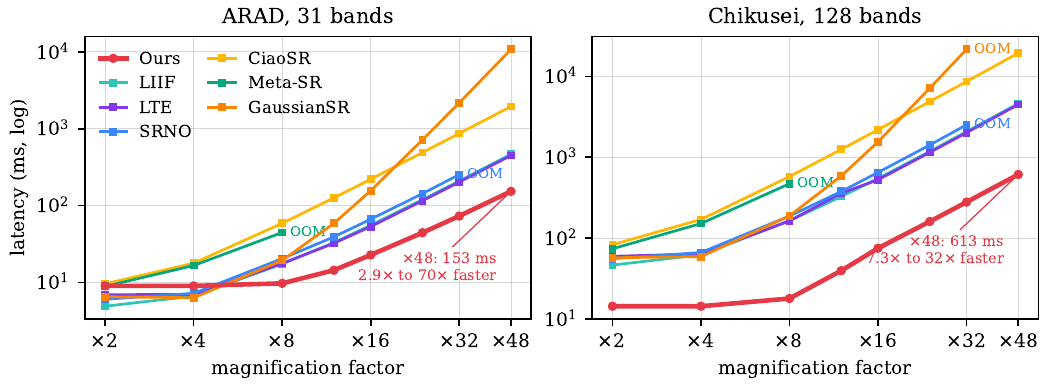}
\caption{Latency against magnification factor on one RTX~4090D (24~GB). The same $32\times32$ low-resolution input is magnified by $\times2$ to $\times48$ (outputs of $64$ to $1536$ pixels), left on the $31$-band ARAD and right on the $128$-band Chikusei, log-log axes. The x-axis is the magnification factor of a single output and every point is one independent inference, not the accumulated time of rendering many scales from one encoding. A baseline stops where all three inputs run out of memory (OOM). The annotation gives the latency of \ours{} at $\times48$ and its ratio to the fastest and the slowest baseline that reach this factor.}
\label{fig:scale_time}
\end{figure}

\paragraph{Latency on an A40.} Table~\ref{tab:efficiency} reports latency on one NVIDIA A40; all seven datasets, $\times8$ and peak memory are in Table~\ref{tab:efficiency_full}. On the $31$-band ARAD at $\times4$, every method takes between $9$ and $33$~ms and \ours{} takes $11.8$~ms; from $\times8$ on, \ours{} is the fastest, and at $\times16$ it is $2.7$ times faster than the fastest baseline. With more than $31$ bands, the output head of a baseline is tied to $31$ bands, so it runs the network $7$ to $11$ times over sliding windows of bands and its latency grows in proportion; \ours{} runs once, its latency at $\times4$ rises only from $11.8$~ms at $31$ bands to $22.3$~ms at $145$ bands, and it is $3.6$ to $8.5$ times faster than the strongest baseline, SRNO, and up to $36.5$ times faster than the slowest. At $\times4$, \ours{} has the lowest peak memory of all methods. Figure~\ref{fig:scale_time} pushes the factor to $\times48$: the latency of \ours{} grows with the number of output pixels but with the lowest slope, reaching $613$~ms at $\times48$ on $128$ bands, $7.3$ to $32$ times faster than the baselines that reach this factor; Meta-SR runs out of memory from $\times12$ on, and SRNO and GaussianSR at $\times48$. The time column of Table~\ref{tab:main} and the time and memory of Table~\ref{tab:ablation} are measured on one RTX~4090D with a batch size of 1 in FP32, on the first test image of each dataset, as the median of 12 synchronized runs after three warm-ups; Table~\ref{tab:main} averages them over the twelve factors.

\begin{table}[!htbp]
\centering
\caption{Inference latency in ms on one NVIDIA A40 at $\times$4 and $\times$16 on four band counts, batch size 1, FP32, $32{\times}32$ input. \textbf{Bold}: lowest.}
\label{tab:efficiency}
\footnotesize\renewcommand{\arraystretch}{1.12}\setlength{\tabcolsep}{3.8pt}
\begin{adjustbox}{max width=\linewidth,center}
\begin{tabular}{@{}l c cc cc cc cc @{}}
\toprule
\multirow{2}{*}{Method} & \multirow{2}{*}{Params (M)} & \multicolumn{2}{c}{ARAD (31)} & \multicolumn{2}{c}{Pavia~U (103)} & \multicolumn{2}{c}{Chikusei (128)} & \multicolumn{2}{c}{Botswana (145)} \\
\cmidrule(lr){3-4}\cmidrule(lr){5-6}\cmidrule(lr){7-8}\cmidrule(lr){9-10}
 & & $\times$4 & $\times$16 & $\times$4 & $\times$16 & $\times$4 & $\times$16 & $\times$4 & $\times$16 \\
\midrule
LIIF & 2.03 & 11.6 & 120.2 & 76.8 & 831.1 & 109.5 & 1187.1 & 120.5 & 1304.1 \\
LTE & 2.18 & 12.4 & 124.1 & 80.3 & 853.9 & 114.1 & 1224.1 & 127.3 & 1342.8 \\
Meta-SR & 6.27 & 23.8 & 271.1 & 171.0 & 1914.3 & 237.0 & 2765.6 & 264.6 & 3024.7 \\
CiaoSR & 3.11 & 33.1 & 450.4 & 227.0 & 3147.8 & 322.4 & 4489.7 & 354.1 & 4934.4 \\
SRNO & 2.48 & 12.3 & 106.0 & 79.0 & 727.9 & 112.1 & 1039.7 & 123.9 & 1142.5 \\
GaussianSR & 6.97 & \textbf{8.9} & 296.7 & 58.3 & 2061.8 & 82.1 & 2939.3 & 90.3 & 3238.0 \\
\midrule
\cellcolor{oursrow}\ours{} & \cellcolor{oursrow}0.538 & \cellcolor{oursrow}11.8 & \cellcolor{oursrow}\textbf{39.1} & \cellcolor{oursrow}\textbf{22.1} & \cellcolor{oursrow}\textbf{148.7} & \cellcolor{oursrow}\textbf{21.4} & \cellcolor{oursrow}\textbf{129.1} & \cellcolor{oursrow}\textbf{22.3} & \cellcolor{oursrow}\textbf{135.1} \\
\midrule
Forward passes of a baseline & & 1 & 1 & 7 & 7 & 10 & 10 & 11 & 11 \\
\bottomrule
\end{tabular}
\end{adjustbox}
\end{table}

Table~\ref{tab:efficiency_full} is the full version of Table~\ref{tab:efficiency}: latency and peak memory on all seven datasets at $\times4$, $\times8$ and $\times16$, on the same A40 with the same timing discipline. Latency depends only on the input size and the band count, so the three $31$-band datasets take the same time, and so do Pavia~C and Pavia~U with $102$ and $103$ bands.

\begin{table}[!htbp]
\centering
\caption{Full version of Table~\ref{tab:efficiency}: latency in ms / peak allocated memory in MiB on one NVIDIA A40, batch size 1, FP32, from a $32{\times}32$ low-resolution input (median of 12 synchronized runs, averaged over three inputs), for all seven datasets at $\times$4, $\times$8 and $\times$16. On inputs with more than $31$ bands a baseline runs its $31$-band model on sliding windows of bands, which takes the number of forward passes in the last row; \ours{} runs once. No configuration ran out of memory on the 46~GB card. \textbf{Bold}: lowest latency. Params counts the source model.}
\label{tab:efficiency_full}
\scriptsize\renewcommand{\arraystretch}{1.12}\setlength{\tabcolsep}{2.5pt}
\begin{adjustbox}{max width=\linewidth,center}
\begin{tabular}{@{}l c c *{7}{c}@{}}
\toprule
Method & Params & Factor & ARAD & CAVE & Harvard & Pavia~U & Pavia~C & Chikusei & Botswana \\
 & (M) & & (31) & (31) & (31) & (103) & (102) & (128) & (145) \\
\midrule
\multirow{3}{*}{LIIF} & \multirow{3}{*}{2.030} & $\times$4 & 11.6\,/\,134 & 11.6\,/\,134 & 11.6\,/\,134 & 76.8\,/\,141 & 77.1\,/\,141 & 109.5\,/\,143 & 120.5\,/\,144 \\
 &  & $\times$8 & 34.0\,/\,141 & 34.1\,/\,141 & 34.1\,/\,141 & 229.4\,/\,167 & 232.7\,/\,167 & 328.1\,/\,174 & 360.3\,/\,178 \\
 &  & $\times$16 & 120.2\,/\,167 & 120.2\,/\,167 & 121.0\,/\,167 & 831.1\,/\,272 & 829.0\,/\,270 & 1187.1\,/\,296 & 1304.1\,/\,314 \\
\midrule
\multirow{3}{*}{LTE} & \multirow{3}{*}{2.180} & $\times$4 & 12.4\,/\,123 & 12.4\,/\,123 & 12.5\,/\,123 & 80.3\,/\,130 & 80.9\,/\,130 & 114.1\,/\,132 & 127.3\,/\,133 \\
 &  & $\times$8 & 35.2\,/\,130 & 38.9\,/\,130 & 35.5\,/\,130 & 237.1\,/\,156 & 236.6\,/\,156 & 337.3\,/\,163 & 370.6\,/\,167 \\
 &  & $\times$16 & 124.1\,/\,156 & 125.5\,/\,156 & 124.2\,/\,156 & 853.9\,/\,261 & 857.9\,/\,259 & 1224.1\,/\,285 & 1342.8\,/\,312 \\
\midrule
\multirow{3}{*}{Meta-SR} & \multirow{3}{*}{6.270} & $\times$4 & 23.8\,/\,2305 & 24.3\,/\,2305 & 24.5\,/\,2305 & 171.0\,/\,2313 & 168.7\,/\,2313 & 237.0\,/\,2315 & 264.6\,/\,2316 \\
 &  & $\times$8 & 74.9\,/\,9122 & 74.6\,/\,9122 & 75.5\,/\,9122 & 553.1\,/\,9149 & 546.0\,/\,9149 & 795.9\,/\,9155 & 859.4\,/\,9160 \\
 &  & $\times$16 & 271.1\,/\,36388 & 271.3\,/\,36388 & 274.3\,/\,36388 & 1914.3\,/\,36493 & 1932.1\,/\,36491 & 2765.6\,/\,36518 & 3024.7\,/\,36536 \\
\midrule
\multirow{3}{*}{CiaoSR} & \multirow{3}{*}{3.110} & $\times$4 & 33.1\,/\,874 & 33.2\,/\,874 & 33.1\,/\,874 & 227.0\,/\,881 & 226.8\,/\,881 & 322.4\,/\,883 & 354.1\,/\,884 \\
 &  & $\times$8 & 125.1\,/\,881 & 119.5\,/\,881 & 117.5\,/\,881 & 813.0\,/\,908 & 812.4\,/\,908 & 1159.7\,/\,914 & 1271.2\,/\,918 \\
 &  & $\times$16 & 450.4\,/\,911 & 451.1\,/\,911 & 450.4\,/\,911 & 3147.8\,/\,1016 & 3153.5\,/\,1014 & 4489.7\,/\,1041 & 4934.4\,/\,1059 \\
\midrule
\multirow{3}{*}{SRNO} & \multirow{3}{*}{2.480} & $\times$4 & 12.3\,/\,264 & 12.2\,/\,264 & 12.3\,/\,264 & 79.0\,/\,271 & 79.4\,/\,271 & 112.1\,/\,273 & 123.9\,/\,274 \\
 &  & $\times$8 & 31.0\,/\,1001 & 31.2\,/\,1001 & 31.1\,/\,1001 & 205.3\,/\,1028 & 205.7\,/\,1028 & 292.2\,/\,1035 & 320.9\,/\,1039 \\
 &  & $\times$16 & 106.0\,/\,3951 & 105.9\,/\,3951 & 105.8\,/\,3951 & 727.9\,/\,4056 & 727.4\,/\,4054 & 1039.7\,/\,4080 & 1142.5\,/\,4099 \\
\midrule
\multirow{3}{*}{GaussianSR} & \multirow{3}{*}{6.970} & $\times$4 & \textbf{8.9}\,/\,273 & \textbf{9.0}\,/\,273 & \textbf{9.0}\,/\,273 & 58.3\,/\,281 & 58.4\,/\,281 & 82.1\,/\,284 & 90.3\,/\,284 \\
 &  & $\times$8 & 33.7\,/\,688 & 34.2\,/\,688 & 33.2\,/\,688 & 231.7\,/\,715 & 227.1\,/\,715 & 328.0\,/\,721 & 358.9\,/\,726 \\
 &  & $\times$16 & 296.7\,/\,2345 & 296.1\,/\,2345 & 295.8\,/\,2345 & 2061.8\,/\,2449 & 2062.5\,/\,2447 & 2939.3\,/\,2474 & 3238.0\,/\,2492 \\
\midrule
\cellcolor{oursrow} & \cellcolor{oursrow} & \cellcolor{oursrow}$\times$4 & \cellcolor{oursrow}11.8\,/\,66 & \cellcolor{oursrow}12.5\,/\,66 & \cellcolor{oursrow}12.6\,/\,57 & \cellcolor{oursrow}\textbf{22.1}\,/\,110 & \cellcolor{oursrow}\textbf{21.2}\,/\,102 & \cellcolor{oursrow}\textbf{21.4}\,/\,88 & \cellcolor{oursrow}\textbf{22.3}\,/\,89 \\
\cellcolor{oursrow} & \cellcolor{oursrow} & \cellcolor{oursrow}$\times$8 & \cellcolor{oursrow}\textbf{17.6}\,/\,209 & \cellcolor{oursrow}\textbf{17.8}\,/\,206 & \cellcolor{oursrow}\textbf{16.3}\,/\,169 & \cellcolor{oursrow}\textbf{48.2}\,/\,355 & \cellcolor{oursrow}\textbf{46.7}\,/\,347 & \cellcolor{oursrow}\textbf{42.8}\,/\,259 & \cellcolor{oursrow}\textbf{45.4}\,/\,254 \\
\cellcolor{oursrow}\multirow{-3}{*}{\ours{}} & \cellcolor{oursrow}\multirow{-3}{*}{0.538} & \cellcolor{oursrow}$\times$16 & \cellcolor{oursrow}\textbf{39.1}\,/\,764 & \cellcolor{oursrow}\textbf{39.2}\,/\,761 & \cellcolor{oursrow}\textbf{33.6}\,/\,623 & \cellcolor{oursrow}\textbf{148.7}\,/\,1317 & \cellcolor{oursrow}\textbf{147.8}\,/\,1319 & \cellcolor{oursrow}\textbf{129.1}\,/\,959 & \cellcolor{oursrow}\textbf{135.1}\,/\,927 \\
\midrule
\multicolumn{3}{@{}l}{Forward passes of a baseline} & 1 & 1 & 1 & 7 & 7 & 10 & 11 \\
\bottomrule
\end{tabular}
\end{adjustbox}
\end{table}

\end{document}